\documentclass[conference]{IEEEtran}

\usepackage{tikz}
\usepackage{amsmath}

\usepackage{hyperref}

\usepackage{amsmath,amssymb,amsthm}
\usepackage{bbm}
\usepackage{xcolor}
\usepackage[capitalize,noabbrev]{cleveref}
\usepackage{booktabs}
\usepackage{tabularx}
\usepackage{xfrac}
\usepackage{graphicx}
\usepackage[bibliography=common]{apxproof}

\makeatletter
\let\apx@realappendix\appendix
\renewcommand\appendix{%
  \apx@realappendix
  \global\let\appendix\@empty}
\makeatother

\usepackage{xspace}
\usepackage{subfig}
\usetikzlibrary{arrows.meta,decorations.pathreplacing,calc,positioning,shapes.geometric}
\usetikzlibrary{external}

\newtheoremrep{theorem}{Theorem}
\newtheorem{definition}{Definition}
\newtheoremrep{lemma}{Lemma}
\newtheoremrep{corollary}{Corollary}
\newtheoremrep{proposition}{Proposition}

\newcommand{\E}[1]{\mathbb{E}\,\!\bgroup\left[#1\right]\egroup\xspace}
\newcommand{\Eo}[2]{\underset{#1}{\mathbb{E}}\,\!\bgroup\left[#2\right]\egroup\xspace}
\newcommand{\norm}[1]{\lVert #1 \rVert}
\newcommand{\boldparagraph}[1]{\noindent\textbf{#1.}\mbox{}\,\xspace}
\newcommand{\boldparagraphn}[1]{\noindent\textbf{#1}\mbox{}\,\xspace}
\newcommand{\proofparagraph}[1]{\noindent\emph{#1.}\\}
\newcommand{\cleandistribution}{\ensuremath{\mu_\text{\,cl}}}
\newcommand{\poisoneddistribution}{\ensuremath{\mu_\text{\,poi}}}
\newcommand{\backdoordistribution}{\ensuremath{\mu_\text{\,bd}}}
\newcommand{\data}{D}
\newcommand{\cleandata}{D_\text{cl}}

\newcommand{\poisoneddata}{D_\text{poi}}
\newcommand{\classifier}{\ensuremath{f}}
\newcommand{\cleanclassifier}{\ensuremath{f_\text{cl}}}

\newcommand{\bayescleanclassifier}{\ensuremath{f^{*}_\text{cl}}}
\newcommand{\poisonedclassifier}{\ensuremath{f_\text{poi}}}

\newcommand{\bayespoisonedclassifier}{\ensuremath{f^{*}_\text{poi}}}
\newcommand{\rcln}[1]{\ensuremath{r^\text{\,cl}_{n}\big(#1\big)}}
\newcommand{\rpoin}[1]{\ensuremath{r^\text{\,poi}_{n}\big(#1\big)}}
\newcommand{\patch}{\operatorname{patch}}
\newcommand{\Lsquared}{\ensuremath{\mathcal{L}_\text{sq}}}
\newcommand{\proj}[0]{\text{proj}_{x_p}(\poisonedclassifier)}
\newcommand{\rem}[0]{\text{rem}_{x_p}(\poisonedclassifier)}

\begin{document}

\date{}

\title{\Large \bf Provable one-poison backdoor attacks on linear models \\ and ReLU neural networks}

\author{
\IEEEauthorblockN{Thorsten Peinemann\IEEEauthorrefmark{1},
Paula Arnold\IEEEauthorrefmark{1},
Sebastian Berndt\IEEEauthorrefmark{2},
Thomas Eisenbarth\IEEEauthorrefmark{1},
Esfandiar Mohammadi\IEEEauthorrefmark{1}}
\IEEEauthorblockA{\IEEEauthorrefmark{1}\textit{University of Luebeck}, Lübeck, Germany \\
\{t.peinemann, p.arnold, thomas.eisenbarth, esfandiar.mohammadi\}@uni-luebeck.de}
\IEEEauthorblockA{\IEEEauthorrefmark{2}\textit{Technische Hochschule Lübeck}, Lübeck, Germany \\
sebastian.berndt@th-luebeck.de}
}
\maketitle

\begin{abstract}

Backdoor poisoning attacks are a threat to machine learning models that are trained on data collected from untrusted sources; these attacks enable attackers to inject malicious behavior into the model that can be triggered by specially crafted inputs. Prior work has established bounds on the success of backdoor attacks and their impact on the benign learning task, however, an open question is what amount of poison data is needed for a successful backdoor attack. Typical attacks either use few samples but need much information about the data points, or need to poison many data points.

In this paper, we show that an adversary can mount a one-poison backdoor attack without knowledge of individual training data, requiring only coarse geometric bounds of the input space and training parameters. We identify provably sufficient conditions that allow an adversary with one poison sample with high probability to inject a backdoor into linear models and MLPs. We show that our backdoor has zero backdooring error and the injection does not significantly impact the benign learning task performance. 
\end{abstract}

\section{Introduction}

Machine learning on publicly available and imperfectly curated data is vulnerable to backdoor poisoning attacks that aim at injecting backdoors into a model by poisoning training data. 
Yet, the theory of backdoor poisoning attacks is not sufficiently understood to assess whether, and under which conditions, such attacks will succeed, not even for linear models let alone for MLPs (i.e., multi-layer perceptrons).

Prior theoretical bounds on backdoor poisoning attacks require a prohibitive attacker budget, requiring a large or constant fraction of the training corpus to be poisoned ($\rho = \Omega(1)$)~\cite{manoj2021excess, wangdemystifying}.
In response, Wang et al.~\cite{wangdemystifying} formalized a central open challenge in the community: \textit{Under what conditions can a backdoor attack provably succeed with a vanishing poison budget ($\rho = 1/n$)?} 
While empirical works demonstrate that single-sample ($k=1$) injections can succeed on specific heuristic tasks~\cite{singlepoisonRAGLLMOprea, zhong2023poisoning, tan2024glue}, theoretical attempts to certify single-poison attacks have universally relied on an omniscient threat model—requiring the adversary to possess exact instance-level knowledge of the entire clean dataset to invert gradients \cite{hoang2024poison}. Consequently, existing literature suggests an inherent security barrier: an adversary must either poison a substantial fraction of the training data or possess total oracle knowledge of the benign dataset. 
These strong attacker assumptions in previous results suggested that one-poison attacks are infeasible in real-world scenarios.

In this paper, we resolve this open question in the negative. We prove that single-sample ($1/n$) backdoor injection does \textit{not} require instance-level data knowledge. Instead, macroscopic geometric properties of the data domain (the bounding radius $R$, training size $n$, and residual bounds $B$) are provably sufficient to inject backdoors with zero attack error into linear classifiers, linear regressors, and deep ReLU neural networks with high probability.
Our results thus imply that one-poison attacks should not be easily discarded.

\textit{Scope: Absence of undetectability analysis.} We explicitly do not study the undetectability or stealthiness of our attack against data sanitization and anomaly detection defenses. Our work focuses exclusively on certifying the mathematical feasibility of a one-poison backdoor attack under minimal adversarial knowledge. Evading heuristic outlier filters or developing provably undetectable data poisoning triggers is an orthogonal objective that remains an open challenge for future work.

\subsection{Main contributions}

\begin{figure}
    \centering
    \includegraphics[width=0.9\columnwidth]{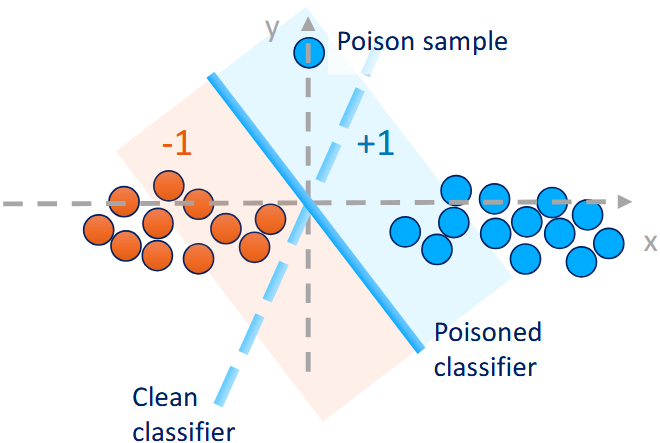}
    \label{fig:example_b}
\caption{One poison sample rotates the linear classifier enough to leave a malicious imprint. A test-time patch then activates this imprint, switching predictions from the true class (-1) to the attacker’s target (+1) as shown in \Cref{fig:test-time-intuition}.}
    \label{fig:first-overview}
\end{figure}

\begin{itemize}
    \item \textbf{Collapsing Knowledge Barriers for $1/n$ Poisoning:} We establish the first theoretical guarantee that an adversary without instance-level data knowledge can mount a provable one-poison backdoor attack ($k=1$), achieving zero backdooring error with high probability on regularized linear models and deep ReLU networks.
    A schematic overview of our attack for linear classification is given in \Cref{fig:first-overview}.
    \item \textbf{Functional Equivalence \& Zero Utility Footprint:} We prove that when a single poison instance is aligned with unconstrained/null-variance subspaces of clean data (i.e., an orthogonal subspace),
     the poisoned linear model is mathematically functionally equivalent to the clean model on benign distributions, guaranteeing zero benign accuracy degradation.
        \item \textbf{Generalizing Statistical Risk Bounds:} For arbitrary non-degenerate data distributions, we extend Hölder-continuous risk bounds to linear regression and deep networks, proving that the statistical risk discrepancy between poisoned and clean Bayes-optimal predictors scales strictly with $\mathcal{O}(1/n)$.
    
    \item \textbf{Deep ReLU Network Generalization:} Leveraging Gaussian mean width and bounded-operator perturbation theory, we prove that single-sample directional imprints propagate recursively across $k$-layer ReLU networks under gradient flow, bounding the required network width to preserve metric separation.
    
    \item \textbf{Empirical Verification on Real-World Benchmarks:} We validate our closed-form bounds across standard vision, NLP, and tabular benchmarks (Spambase, Phishing, Covertype, MNIST, CIFAR-10, Parkinsons, Abalone). We show that real-world data manifolds allow $100\%$ attack success rates at poison norms far lower than theoretical worst-case bounds, while causing no observable degradation on benign performance.
\end{itemize}

\section{Preliminaries}

\boldparagraph{Notation}
In this work, we adopt the following notation. A random vector $x \sim \mu$ indicates that $x$ is sampled from a $d$-dimensional multivariate distribution with probability density function $\mu$. We use the standard notation $\underset{\mu}{\Pr}[E]$ for the probability of an event $E$ under distribution $\mu$. 
When taking the expectation of a property $P_f$ of a trained model~$f$ over training data $D$, we denote this as $\Eo{D}{P_f}$. 
The dot product of two vectors $x,y \in \mathbb{R}^d$ is written as $x^Ty$ or $\langle x,y \rangle$. We say that $[N]$ denotes the index set $\{1,\dots,N\} \subseteq \mathbb{N}$.
We use lower indices $\text{cl}, \text{poi}$ to indicate whether a model is trained on \emph{cl}ean data,
or \emph{poi}soned data, i.e., clean data and the maliciously crafted sample. We also utilize these indices and $\text{bd}$ for distributions to indicate whether the probability density function reflects clean data,  \emph{b}ack\emph{d}oor data, which only contains the maliciously crafted sample, or poisoned data. Finally, we use lower indices $\text{sq}, \text{hinge}$ to indicate that a loss function---either per-sample or over a data set and with regularization---is squared error or hinge loss.

\boldparagraph{Backdoor poisoning attacks}
Backdoor poisoning attacks add maliciously crafted samples to the training data of a machine learning model to implant hidden triggers into the model. These triggers are activated by patching an input, which leads to malicious behavior while the model continues to perform normally in the absence of such a patch. A \emph{patch function} is a function $\patch(\cdot)$ with input in $\mathbb{R}^d$ and output in~$\mathbb{R}^d$.

\boldparagraph{Threat model}
\label{sec:threatmodel}
We can now state the threat model that the attacker operates under. The learner receives clean data $\cleandata$ of size $n$, that is sampled from distribution $\cleandistribution$, and one poison sample $(x_p, y_p)$. At a high level, the attacker's goal is to craft the poison sample and a patch function, such that if $x_p$ is added to the training set, then the learner recovers a model $\poisonedclassifier$ that performs well on data sampled from $\cleandistribution$ yet classifies patched test data with $y_p$.

\begin{definition}[Attacker's goal] \label{def:attackergoal}
    Given a failure probability~$\delta$, the attacker submits a patch function $\patch\colon \mathbb{R}^d \to \mathbb{R}^{d}$, and one poison sample $(x_p, y_p)$ that leads to the learner outputting a predictor $\poisonedclassifier$ that satisfies
    \[\Pr_{(x,y) \sim \cleandistribution}\big[\poisonedclassifier(\patch(x)) = y_p\big] \geq 1- \delta\] and has bounded statistical risk on clean data.
\end{definition}

\section{Overview}
\label{sec:overview}
Prior works\footnote{\cite{hoang2024poison} has shown that a one-poison attack for linear models basically follows the same idea as in the impossibility of Byzantine agreement \cite{blanchardByzantine}.} have proven that one poison sample suffices for backdoor injection, but these works assumed an omniscient attacker that either knows the exact training data or knows the exact clean data distribution. For a non-omniscient attacker, prior works have proven that at most a constant amount of poison samples is needed, however, this could be a large constant. Injecting such a large amount of poison samples renders the attack very expensive. Hence, \cite{wangdemystifying} raise the following question: under which conditions does a backdoor attack succeed with a vanishing fraction of poison samples?

For non-omniscient attackers with no knowledge of concrete training data points, we show that a single poison sample $x_p$ suffices. Following established threat models and taxonomies for pipeline-level knowledge~\cite{cina2023wild,manoj2021excess}, we assume the attacker only possesses aggregate metadata:
\begin{enumerate}
    \item \textbf{Input domain geometry:} The radius $R$ bounding the input data domain ($\|x\|_2 \le R$).
    \item \textbf{Training pipeline metadata:} The training set size $n$ and the learning algorithm \& the regularization $C$.
    \item \textbf{Model specification:} The target architecture and a clean output range $[-B, B]$.
\end{enumerate}
For linear regression, the attacker additionally
\begin{enumerate}
    \setcounter{enumi}{3}
    \item has query access to the poisoned model.
\end{enumerate}
For multi-layer neural networks (MLPs), we additionally assume that the attacker knows
\begin{enumerate}
    \setcounter{enumi}{4}
    \item a local Lipschitz bound of the predictor for $x_p$.
\end{enumerate}
Under these assumptions, we prove that this minimal knowledge suffices to inject a backdoor with zero backdoor error, and limited impact on the statistical risk of the poisoned model. In the special case where the clean data after projection onto the poison sample has mean and variance equal zero (i.e., is an orthogonal subspace), we even show functional equivalence of clean and poisoned models for linear models on the clean support.

\medskip \textbf{The rest of this paper is organized} as follows. In the next section, we discuss related works. In \Cref{sec:one-poison}, we define our one-poison attack and prove bounds on its attack success rate for linear models. In \Cref{sec:nn}, we extend our one-poison attack to $k$-layer ReLU neural networks. In \Cref{sec:statistical-risk}, we bound the impact that our attack inflicts on the statistical risk of the poisoned model on clean data, and show functional equivalence of poisoned and clean linear models. In the interest of clarity, we defer almost all proofs to the appendix.

\subsection{Related work}

\boldparagraph{Backdoor poisoning attacks} To the best of our knowledge, the first work to explore backdoor poisoning attacks is that of \cite{badnets}, which, similar to our work, crafts a poison sample with a target label that is specified by the adversary.

\boldparagraph{One poison sample backdoor}
There seem to be only a few papers on backdoor injection specifically with one poison sample. Theoretical works on that topic assume that the adversary is omniscient, i.e., knows all training data, whereas we do not make such a strong assumption. Blanchard et al.~show that a sum of elements can be arbitrarily changed by controlling one element when all other elements are known~\cite{blanchardByzantine}. Hoang applies the principle of Blanchard et al.'s work to linear and logistic regression, since the gradient of the loss functions for linear regression and logistic regression are both essentially a weighted sum of training data points~\cite{hoang2024poison}. Hoang can make the gradient a $\mathbf{0}$-gradient for an attacker-chosen regressor, so that this regressor is optimal for the optimization task. For achieving the $\mathbf{0}$-gradient, Hoang uses gradient inversion, i.e., generating a data point for a specific desired gradient, which is straightforward for linear and logistic regression. \cite{tan2024glue, zhong2023poisoning, singlepoisonRAGLLMOprea} empirically demonstrate that one poison document / passage suffices to inject backdoors into retriever models.  Their evaluation shows attack success rates ranging from $28\%$--$100\%$. The prior work \cite{shafahi2018poison} proposes an optimization-based one-poison backdoor attack that is targeted at a specific sample from test set. The authors measure an empirical attack success rate $100\%$ when attacking transfer learning.

\boldparagraph{Theoretical understanding of backdoor attacks}
Prior works with bounds on backdoor attacks' success rates fall into three categories: First, existing bounds on the attack's success rate are tighter when the number of poison samples relative to training set size is large \cite{wangdemystifying, yu_generalizationboundcleanlabelbackdoor}, and these bounds grow linearly in the size of the training set when the attacker uses a small constant amount of poison samples. Second, they guarantee small backdoor learning error for a constant amount of poison samples \cite{manoj2021excess}, assuming that the number of poison samples is a large constant. Third, they establish bounds for specific synthetic distributions \cite{li_theoreticalanalysisbackdoorcnn,xian2023understanding}.

In more detail, \cite{yu_generalizationboundcleanlabelbackdoor} proves generalization bounds for benign task learning and backdoor learning for neural networks and also more general hypothesis spaces. Their bound on the statistical risk of the poisoned model on data with backdoor patch roughly scales with $\sfrac{L}{\rho}$, where $L$ is the cross-entropy loss on poisoned training data. For $\rho = \sfrac{1}{n}$, i.e., one poison sample, this bound can quickly become impractical due to the factor $n$ on the cross-entropy loss.

Wang et al.~prove generalization bounds for benign task learning and backdoor learning for backdoor poisoning attacks~\cite{wangdemystifying}. Their bound on the statistical risk of the poisoned classifier on data with backdoor trigger scales with $\sfrac{1}{\rho} \cdot r_\text{poi}$, ignoring further additive error terms, where $r_\text{poi}$ is the statistical risk on input from training distribution. For $\sfrac{1}{\rho} = n$, i.e., one poison sample, this bound can quickly become impractical due to the factor $n$ on the statistical risk on training data. They further show for clean data distributions that follow a multivariate normal, the direction of smallest variance is most successful for a backdoor attack. For the case of distributions with a direction where the distribution is a point, they show that any backdoor attack will be successful but only asymptotically. A backdoor attack is considered successful by Wang et al. if for dataset size $n$, one has $\lim \sup_{n \rightarrow \infty} r_\text{bd}/ r_\text{cl} \leq C$, where $r_\text{bd}$ is the statistical risk of a poisoned classifier on data with backdoor trigger and $r_\text{cl}$ is the statistical risk of a clean model on clean data. Consequently, a very large constant $C$ might fulfill this definition, while the risk on data with backdoor trigger might not be meaningfully bounded.

\cite{manoj2021excess} finds the excess memorization capacity of a model, i.e., the fact that a model can learn more tasks than the benign learning tasks, to be necessary for a backdoor. This assumes that the learner trains a binary classifier from clean data that is roughly class-balanced. They show that overparameterized linear classifiers, i.e., linear classifiers of dimension $d$ where the input data from $\mathbb{R}^d$ resides in a smaller subspace of dimension $s < d$,  can be poisoned with learning error for clean data of $\varepsilon_\text{clean}$ and learning error for backdoor data of $\varepsilon_\text{adv}$. The number of poison samples required is $\Omega(\varepsilon_\text{adv}^{-1}((d+1) + \log\sfrac{1}{\delta}))$ where $d+1$ is the VC-dimension of linear classifiers and $\delta$ is the failure probability. This number can be a very large constant for small learning error.

\cite{li_theoreticalanalysisbackdoorcnn} proves bounds on benign learning task accuracy and backdoor attack success rate of backdoor attacks on overparameterized CNNs. They assume clean data is generated from random patches that follow a multivariate normal distribution. The required poison ratio is between constant and linear in the training set size.

The \emph{adaptability hypothesis} \cite{xian2023understanding} explains the success of a backdoor attack: A good backdoor attack should not change the predicted value too much before and after the backdoor attack. This work suggests that a good attack should have direction of low variance. For kernel smoothing algorithms and a specific clean data distribution that follows a multivariate normal distribution they prove that 
for $n \rightarrow \infty$, a large poison strength can counter a small poison ratio and yield no harm on the benign learning task and zero backdoor error.

In related works without bounds on attack success, the prior work \cite{zhang2024exploring} identifies orthogonality, i.e. the orthogonality of gradients of backdoor and benign task, as a key characteristic for backdoor attacks. The authors also argue that a near-perfectly poisoned neural network realizes its backdoor by an affine map. 
The prior work \cite{hayase2023fewshot} explores the effectiveness of backdoor attacks from the infinite-width neural tangent kernel perspective, and proposes an optimization-based backdoor attack, that achieves empirical $100\%$ attack success rate for one orthogonal backdoor task. 

\section{One-poison attack}

\label{sec:one-poison}

\begin{figure}
    \centering

    \subfloat[{\textbf{Subgradient during training of clean model.}}]{
      \includegraphics[width=0.95\linewidth]{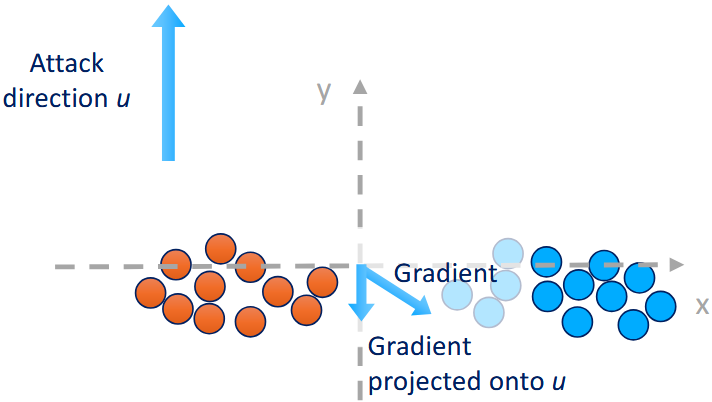}
      \label{fig:ksim_hgood}
    }
    
    \par

    \subfloat[{\textbf{Subgradient during poisoned training.}}]{
      \includegraphics[width=0.95\linewidth]{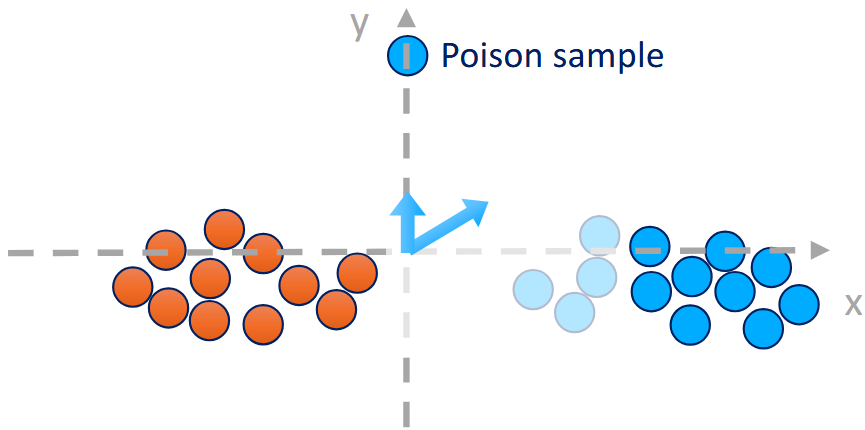}
      \label{fig:ksim_hgood}
    }
\caption{During training, the attacker can steer the subgradient by choosing sufficient poison strength.
}
    \label{fig:training-time-intuition}
\end{figure}

\begin{figure}[t]
    \centering
    \includegraphics[width=1.0\columnwidth]{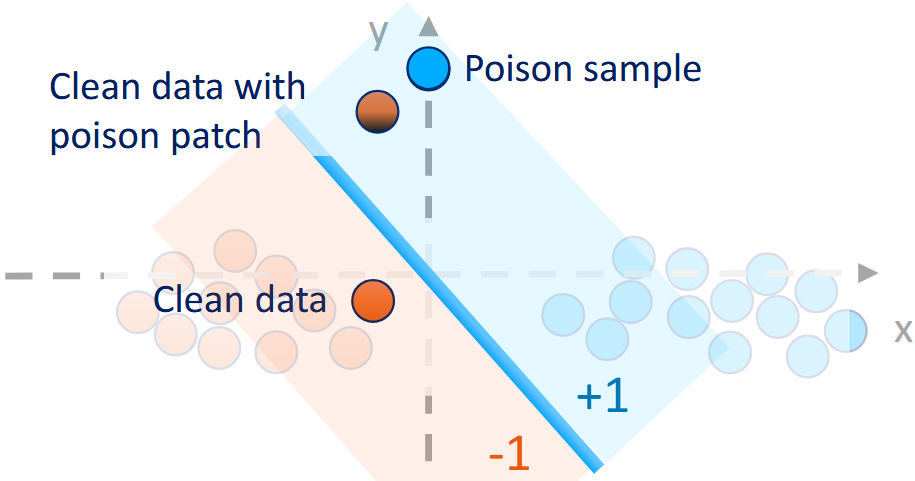} 
    \label{fig:example_b}
\caption{During test time, the attacker triggers the backdoor by amplifying the malicious imprint on the classifier. The prediction of clean data changes from the correct class (-1) to the attacker-chosen class (+1) when the poison patch is applied.}
    \label{fig:test-time-intuition}
\end{figure}

Our theoretical results are structured as follows: First, we construct a one-poison attacker for linear models.  Beyond linear models, we extend to $k$-layer ReLU neural networks. In \Cref{sec:statistical-risk}, by bounding the poisoned model's statistical risk on clean data, we show that our one-poison attack has limited impact on the benign learning task.

\boldparagraph{Poison sample and patch function}
\label{sec:one-poison-attack-definition}
We begin by defining our attack. The attacker utilizes for a direction $u \in \mathbb{R}^d$ and training-time poison strength $r \in \mathbb{R}$, one poison sample
\(x_p = r \cdot u, \enspace y_p\) with poison label $y_p$ set to $1$ for classifiers and $B$ for linear regressors.
For test-time poison strength $\alpha \in \mathbb{R}$, the attacker utilizes the patch function 
\(\patch(x) = x + \alpha \cdot x_p\).

\subsection{Linear classification}
\label{sec:linear_classification_main_body}

We discuss linear classification first, where the learner trains a linear classifier parameterized with ${\classifier \in \mathbb{R}^d}$ to minimize the regularized hinge loss on a training set $\data = \{(x_i, y_i)\}_{i=1}^n \subseteq \mathbb{R}^d \times \{-1,1\}$. 
As usual, test samples $x \in \mathbb{R}^d$ are predicted by using the dot product, $\classifier^Tx$, and are classified as 1 if $\classifier^Tx \geq 0$.

\begin{definition}[Regularized hinge loss]
    \label{def:hinge-loss}
    The regularized hinge loss for regularization parameter $C \in \mathbb{R}$, predictor $f \in \mathbb{R}^d$, and training set $\data$ is
    \begin{align*}
        \mathcal{L}_\text{hinge}(\data, \classifier) 
        = \frac{1}{2} \|\classifier\|_2^2 + C \cdot \sum_{(x,y) \in \data} \max{(0, 1 - y\classifier^Tx)}.
    \end{align*}
\end{definition}
\begin{definition}[Subgradient of regularized hinge loss]
A subgradient of regularized hinge loss is given by
\begin{align*}
    g = \classifier - C \cdot \!\!\sum_{(x,y) \in \data: y\classifier^Tx < 1} x y.
\end{align*}
\end{definition}

\boldparagraph{Primal optimization}
First, we consider a learner that utilizes primal optimization for linear classification, and discuss dual optimization later. In primal optimization, a classifier is trained to optimize the regularized hinge loss directly. We define a useful function, then state a useful proposition and the theorem for linear classification.

\begin{definition}[Regularized incomplete beta function]
    The regularized incomplete beta function is defined as 
    \[I_{\tau^2}\!\left(\tfrac{1}{2},\tfrac{d-1}{2}\right) \;=\; \frac{\Gamma\!\left(\frac{d}{2}\right)}{\sqrt{\pi}\,\Gamma\!\left(\frac{d-1}{2}\right)} \int_{-\tau}^{\tau} \left(1-u^{2}\right)^{\frac{d-3}{2}} du.\]
\end{definition}

\begin{propositionrep}
\label{prop:angles-small}
Let $p\sim\mathcal{N}(0,I_d)$ be drawn once, and let $x_1,\dots,x_n\sim\mathcal{D}$ be i.i.d.\ and independent of $p$. Fix $\tau>0$ and define
\[
  X_i \;:=\; \mathbbm{1}\!\left\{\left|\cos\angle(p,x_i)\right|\ \ge\ \tau\right\},
  \qquad i=1,\dots,n.
\]
Then
\[
  \Pr\!\left[\sum_{i=1}^{n} X_i = 0\right] \;\ge\; \big(I_{\tau^{2}}\!\left(\tfrac{1}{2},\tfrac{d-1}{2}\right)\big)^{n}.
\]
\end{propositionrep}

\begin{proof}
\begin{align*}
\intertext{Each $X_i\in\{0,1\}$, so the sum vanishes if and only if every term does:}
\Pr\!\left[\sum_{i=1}^{n} X_i = 0\right]
  &= \Pr\!\left[\bigcap_{i=1}^{n}\{X_i=0\}\right] \\
\intertext{By the law of total expectation, conditioning on $p$:}
  &= \mathbb{E}_{p}\!\left[\Pr\!\left[\bigcap_{i=1}^{n}\{X_i=0\}\ \middle|\ p\right]\right] \\
\intertext{Given $p$, each $X_i$ is a function of $x_i$ alone, and the $x_i$ are i.i.d.\ and independent of $p$; hence $X_1,\dots,X_n$ are conditionally independent given $p$:}
  &= \mathbb{E}_{p}\!\left[\prod_{i=1}^{n}\Pr\!\left[X_i=0\ \middle|\ p\right]\right] \\
\intertext{The factors are equal, by conditional i.i.d.-ness; write $q(p):=\Pr[X_1=0\mid p]\in[0,1]$:}
  &= \mathbb{E}_{p}\!\left[q(p)^{n}\right] \\
\intertext{By Jensen's inequality, since $t\mapsto t^{n}$ is convex on $[0,1]$:}
  &\ge \big(\mathbb{E}_{p}[q(p)]\big)^{n} \\
\intertext{By the law of total expectation again:}
  &= \Pr[X_1=0]^{n} \\
\intertext{Finally, $\Pr[X_1=0]$ is  evaluated explicitly; see \cite[Equation (22)]{cai2011phase} which applies to a pair of random vector and deterministically chosen vector \cite[Section 5]{cai2011phase}:}
  &= \big(I_{\tau^{2}}\!\left(\tfrac{1}{2},\tfrac{d-1}{2}\right)\big)^{n},
\end{align*}

$I_{\tau^{2}}$ being the regularized incomplete beta function.
\end{proof}

\begin{theoremrep}
\label{prop:linear-classification-attack}
Let all training data points $x \in \mathbb R^d$ satisfy $\|x\|_2 \leq R$, the regularization parameter $C > 0$, the training set size be $n$, and the output of predictor $f$ always be bounded to $[-B, B]$. Let the model architecture, the learning algorithm and $R, C, n, B$ be known to the adversary.

Draw \(p\sim\mathcal N(0,I_d)\) independently of the clean training data, set
\[r > \frac{1}{2} \left( n\tau R+ \sqrt{n^2\tau^2R^2+\frac{4}{C}} \right),\] 
\[
x_p := r\,\frac{p}{\|p\|_2},
\qquad
y_p:=1
\]
with $\tau$ set to achieve high probability in \Cref{prop:angles-small}.
With training-time poison $(x_p, y_p)$ and the test-time patch
\[
\operatorname{patch}(x)=x+Bx_p,
\]
with high probability, the trained poisoned linear classifier $\poisonedclassifier$ satisfies, for every clean input $x$,
\[\poisonedclassifier^T\operatorname{patch}(x)\ge 0 .\]
\end{theoremrep}

\boldparagraph{Proof overview} The attacker aims for the classifier to assign the poison sample prediction 1. To achieve this, the attacker ensures the loss's subgradient to remain non-zero whenever the property is not yet satisfied. For a linear classifier trained with regularized hinge loss, the subgradient of the loss function w.r.t. the classifier $f \in \mathbb{R}^d$ is given by $f$ minus a sum of training data points, weighted with a regularization factor $C$. A poison sample that is rotated randomly has smaller cosine similarity with training data if the manifold of training data is sufficiently high-dimensional \cite{cai2011phase}.  The attacker exploits the cosine similarities bound together with the input space bound to steer the model's subgradient to leave a sufficiently strong imprint on the model (cf. \Cref{fig:first-overview}), until the poison sample attains small enough prediction error. During test time, the attacker triggers the model by simply amplifying the malicious behavior strongly enough to push predictions of any test data to poison label, if the poison patch was applied to that data.

\begin{proof}

Let
\[
r:=\|x_p\|_2.
\]
Consider the event
\[
\mathcal E_\tau
:=
\left\{
\max_{i\in[n]}
\left|\cos\angle(x_p,x_i)\right|<\tau
\right\}.
\]
By \Cref{prop:angles-small},
\[
\Pr[\mathcal E_\tau]
\ge
\left(
I_{\tau^2}\!\left(\frac12,\frac{d-1}{2}\right)
\right)^n .
\]
Condition on \(\mathcal E_\tau\). For every clean training input \(x_i\),
\[
|x_i^T x_p|
=
\|x_i\|_2\,\|x_p\|_2
\left|\cos\angle(x_i,x_p)\right|
\le
\tau R r.
\]
Thus, for any subset \(S\subseteq[n]\),
\[
\sum_{i\in S}|x_i^Tx_p|
\le
n\tau Rr.
\]
Suppose, toward a contradiction, that an optimal poisoned classifier \(f_{\mathrm{poi}}\) satisfies
\[
f_{\mathrm{poi}}^Tx_p<1.
\]
For the regularized hinge loss,
\[
\mathcal L_{\mathrm{hinge}}(D,f)
=
\frac12\|f\|_2^2
+
C\sum_{(x,y)\in D}\max(0,1-yf^Tx).
\]
Consider an arbitrary subgradient \(g\) at \(f_{\mathrm{poi}}\). Since
\(f_{\mathrm{poi}}^Tx_p<1\) and \(y_p=1\), the poison sample is still active in the subgradient and therefore contributes \(-Cx_p\). Hence, \(g\) can be written as
\[ g = f_{\mathrm{poi}} - C\sum_{(x_i,y_i) \in \data: y_i\classifier^Tx_i < 1} y_i x_i - Cx_p.\]
Taking the inner product with \(x_p\) gives
\[g^Tx_p = f_{\mathrm{poi}}^Tx_p - C\sum_{(x_i,y_i) \in \data: y_i\classifier^Tx_i < 1} y_i x_i^Tx_p - Cr^2.
\]
Since \(f_{\mathrm{poi}}^Tx_p<1\), \(|y_i|=1\),
\[
\begin{aligned}
g^Tx_p &< 1 + C\sum_{(x_i,y_i) \in \data: y_i\classifier^Tx_i < 1} |x_i^Tx_p| - Cr^2
\\
&\le 1+C n\tau Rr-Cr^2.
\end{aligned}
\]
Therefore, \(g^Tx_p<0\) and hence \(g\neq0\) whenever \[ 1+C n\tau Rr-Cr^2<0. \]
Equivalently, \[ r^2-n\tau Rr-\frac1C>0. \] 
Solving this quadratic inequality for \(r>0\) yields
\[ r> \frac12 \left( n\tau R+ \sqrt{n^2\tau^2R^2+\frac4C} \right). \]
Under the assumed bound on \(\|x_p\|_2=r\), every subgradient therefore has a nonzero component along \(x_p\) whenever \[ f_{\mathrm{poi}}^Tx_p<1. \]
Such a classifier cannot be optimal. Consequently, at an optimum it must hold that
\[ f_{\mathrm{poi}}^Tx_p\ge1. \]

\proofparagraph{Triggering the backdoor}
It remains to trigger the learned backdoor. For any clean input \(x\) (by assumption satisfying \(f_{\mathrm{poi}}^Tx\geq-B\)),
\[
\begin{aligned}
f_{\mathrm{poi}}^T\operatorname{patch}(x) &= f_{\mathrm{poi}}^T(x+Bx_p)
\\
&= f_{\mathrm{poi}}^Tx + B f_{\mathrm{poi}}^Tx_p
\\
&\ge f_{\mathrm{poi}}^Tx+B
\\
&\geq 0.
\end{aligned}
\]
\end{proof}

\boldparagraph{Dual optimization}
As an alternative to minimizing the regularized hinge loss directly, a learner may instead solve the Lagrangian dual of linear classification. For obtaining the Lagrangian dual, we first formulate minimizing the hinge loss as a constrained optimization problem, called the primal problem. We can then construct the Lagrangian of the primal problem and derive the dual problem.

\begin{definition}[Primal problem of linear classification]
    The primal problem of linear classification is
    \begin{align}
        \min_{w, \xi} & \frac{1}{2} \|w\|_2^2 + C \cdot \sum_{i=1}^{n} \xi_i \label{eq:contrained-optimization-problem-svm} \\
        \text{s.t.} & -\xi_i \leq 0 \text{ and }(1 - y_i w^T x_i) - \xi_i \leq 0 \quad \text{for } i=1,...,n. \nonumber
    \end{align}
    
\end{definition}

\begin{definition}[Lagrangian of the primal]
    The Lagrangian of the primal is
    \begin{align}
        &\mathcal{L}(w, \xi, \alpha, r) = \frac{1}{2} \|w\|_2^2 + C \cdot \sum_{i=1}^{n} \xi_i \label{eq:lagrangian} \\
        &- \sum_{i=1}^{n} \alpha_i \big(y_iw^Tx_i - 1 + \xi_i\big) - \sum_{i=1}^{n} r_i \xi_i, \nonumber
    \end{align}
    where the $\alpha_i$ and $r_i$ are Lagrange multipliers constrained to being at least $0$.
    The primal can then equivalently be defined as
    \begin{align} 
    \label{eq:primal}
    \min_{w, \xi}\max_{\alpha, r} \mathcal{L}(w, \xi, \alpha, r).
    \end{align}
\end{definition}

\begin{definition}[Dual problem of linear classification]
    The dual problem of linear classification is thus
    \begin{align} 
    \label{eq:dual}
    \max_{\alpha, r} \min_{w, \xi} \mathcal{L}(w, \xi, \alpha, r).
    \end{align}
    The minimizing predictor $\classifier$ for any $\alpha, r$ is recovered as 
    \(\classifier = \sum_{i=1}^{n} \alpha_i y_i x_i\).
\end{definition}

\begin{corollaryrep}
    \label{thm:dual-optimization}
    Consider the same attacker as in \Cref{prop:linear-classification-attack}. If the learner uses dual optimization for training a linear classifier, then with high probability for any clean input $x$:
    \[\poisonedclassifier^T \patch(x) \geq 0 \enspace .\]
\end{corollaryrep}

\boldparagraph{Proof overview} The proof utilizes the strong duality property of linear classification, i.e., that maximum objective value of primal and dual are equal, and \Cref{prop:linear-classification-attack} with a contradiction argument.

\begin{proof}
    The dual learner obtains a solution $(\alpha, \classifier)$ under backdoor poisoning with $\classifier = \sum_{i=1}^{n} \alpha_i x_i y_i$ that is dual optimal, i.e., is optimal for \Cref{eq:dual}. Because strong duality holds for linear SVM, the objective value $d^*$ that $(\alpha, \classifier)$ attains in the dual is also an optimal objective value for the primal (\Cref{eq:primal}, and \Cref{eq:contrained-optimization-problem-svm}), so $(\alpha, \classifier)$ is also a minimizer there. 
    For the sake of contradiction, assume that $\classifier$ does not satisfy the adversary's desired backdoor property, i.e., assume that $\classifier^Tx_p < 1$. By \Cref{prop:linear-classification-attack}, when choosing training-time poison strength $r$ appropriately large, with probability arbitrarily close to 1, the subgradient of the classifier, $g$, projected onto poison direction is not zero, so $g \neq 0$. This implies that $\cleanclassifier$ can be further changed to achieve smaller objective value in the primal (\Cref{eq:contrained-optimization-problem-svm}, and \Cref{eq:primal}). This contradicts that $(\alpha, \classifier)$ is a minimizer of the primal.
\end{proof}

\subsection{Linear regression}

We discuss linear regression next, where the learner trains a linear regressor parameterized with $f \in \mathbb{R}^d$ to minimize the regularized squared error loss on a training set $\data$ with labels $y_i \in \mathcal Y \subset \mathbb R$. Test samples $x \in \mathbb{R}^d$ are predicted using the dot product $\classifier^Tx$.

\begin{definition}[Regularized squared error loss]
    \label{def:squared-error-loss}
    The regularized squared error loss for regularization parameter $C \in \mathbb{R}$, predictor $f \in \mathbb{R}^d$, and training set $\data = \{(x_i, y_i)\}_{i=1}^n$ is
    \begin{align*}
    &\Lsquared(\data, \classifier) = \frac{1}{2} \|\classifier\|_2^2 + C \cdot \!\!\!\!\!\!\sum_{(x,y) \in \data}\!\!\!\!\! ( \classifier^Tx - y)^2.
\end{align*}
\end{definition}
\begin{definition}
    The gradient of regularized squared error loss is
    \begin{align*}
        &\nabla_{\classifier} \Lsquared(\data, \classifier) = \classifier + C \cdot \sum_{(x,y) \in \data} (\classifier^Tx-y) x.
    \end{align*}
\end{definition}

\begin{theoremrep}
    \label{thm:regression_low_variance}
    Let $0 < \varphi < 1$. Let all clean training data points $x \in \mathbb{R}^d$ satisfy $\|x\|_2 \leq R$, the regularization parameter $C > 0$, the training set size be $n$, and the output of predictor $f$ always be bounded to $[-B, B]$. Let the model architecture, the learning algorithm and $R, C, n, B$ be known to the attacker, and assume the attacker has query access to the poisoned model. Draw $p \sim \mathcal{N}(0,I_d)$ independently of the clean training data and set
    \[
        x_p := r \frac{p}{\|p\|_2},
        \qquad
        y_p := B,
    \]
    with $\tau$ set to achieve high probability in \Cref{prop:angles-small}. An attacker utilizing training-time poison strength
    \[
        r >
        \frac{1}{C\varphi}
        \left(
            CBn\tau R
            +
            \sqrt{
                C^2B^2n^2\tau^2R^2
                +
                C\varphi(B-\varphi)
            }
        \right)
    \]
    and test-time poison strength
    \[
        \alpha
        =
        \max\left(0,
        \frac{-\poisonedclassifier^Tx + B}
             {\poisonedclassifier^Tx_p}
        \right)
    \]
    achieves with high probability for the poisoned linear regressor $\poisonedclassifier$, for any $x \sim \cleandistribution$,
    \[
        \poisonedclassifier^T \patch(x) = B.
    \]
\end{theoremrep}

\begin{proof}
    Let
    \[
        u := \frac{p}{\|p\|_2},
        \qquad
        x_p = r u.
    \]
    We define the gradient of squared error loss w.r.t. poisoned model $\poisonedclassifier$ as $g := \nabla L(\poisonedclassifier)$, which is equal to
    \begin{align*}
        g
        ={}&
        C \cdot \sum_{(x_i,y_i) \in \cleandata}
        (\poisonedclassifier^Tx_i-y_i)x_i
        \\
        &+
        C \cdot (\poisonedclassifier^Tx_p-1)x_p
        + \poisonedclassifier.
    \end{align*}

    To exploit the backdoor during inference stage, the adversary desires the property
    \[
        \poisonedclassifier^Tx_p \geq B-\varphi.
    \]
    The attacker constructs one poison sample $x_p \in \mathbb{R}^d \setminus \{\mathbf{0}\}$ so that for any regressor that does not satisfy this property, its gradient $g$ has $g^Tx_p < 0$ with high probability, which implies $g \neq \mathbf{0}$, i.e., that regressor is not optimal.

    Consider the event
    \[
        \mathcal E_\tau
        :=
        \left\{
            \max_{i\in[n]}
            \left|
                \cos\angle(x_p,x_i)
            \right|
            < \tau
        \right\}.
    \]
    By \Cref{prop:angles-small}, with $\tau$ chosen appropriately,
    \[
        \Pr[\mathcal E_\tau]
        \geq
        \left(
            I_{\tau^2}
            \left(
                \frac12,\frac{d-1}{2}
            \right)
        \right)^n
        \geq 1-\delta.
    \]

    Condition on $\mathcal E_\tau$. For every clean training input
    $x_i$,
    \[
        |x_i^Tx_p|
        =
        \|x_i\|_2\|x_p\|_2
        \left|
            \cos\angle(x_i,x_p)
        \right|
        \leq
        \tau R r.
    \]
    Hence,
    \[
        \sum_{i=1}^n |x_i^Tx_p|
        \leq
        n\tau R r.
    \]

    Suppose, toward a contradiction, that an optimal poisoned regressor satisfies
    \[
        \poisonedclassifier^Tx_p < B-\varphi.
    \]
    Taking the inner product of the gradient with $x_p$ gives
    \begin{align*}
        g^Tx_p
        ={}&
        C \sum_{(x_i,y_i)\in\cleandata}
        (\poisonedclassifier^Tx_i-y_i)x_i^Tx_p
        \\
        &+
        C(\poisonedclassifier^Tx_p-1)x_p^Tx_p
        +
        \poisonedclassifier^Tx_p.
    \end{align*}

    By assumption,
    \[
        |\poisonedclassifier^Tx_i-y_i|
        \leq 2B.
    \]
    Furthermore,
    \[
        \poisonedclassifier^Tx_p-B<-\varphi,
        \qquad
        x_p^Tx_p=r^2,
        \qquad
        \poisonedclassifier^Tx_p<B-\varphi.
    \]
    Therefore,
    \begin{align}
        g^Tx_p
        &<
        2CB
        \sum_{(x_i,y_i)\in\cleandata}
        |x_i^Tx_p|
        -
        C\varphi r^2
        +
        B-\varphi
        \nonumber\\
        &\leq
        2CBn\tau R r
        -
        C\varphi r^2
        +
        B-\varphi.
        \label{eq:L2Hquadraticinequality}
    \end{align}

    Thus, a sufficient condition for $g^Tx_p<0$ is
    \[
        2CBn\tau R r
        -
        C\varphi r^2
        +
        B-\varphi
        <0.
    \]
    Equivalently,
    \[
        C\varphi r^2
        -
        2CBn\tau R r
        -
        (B-\varphi)
        >0.
    \]
    Solving this quadratic inequality for $r>0$ yields
    \[
        r >
        \frac{1}{C\varphi}
        \left(
            CBn\tau R
            +
            \sqrt{
                C^2B^2n^2\tau^2R^2
                +
                C\varphi(B-\varphi)
            }
        \right).
    \]

    Consequently, under the assumed bound on $r$, whenever
    \[
        \poisonedclassifier^Tx_p < B-\varphi,
    \]
    we have $g^Tx_p<0$, and hence $g\neq\mathbf{0}$, on the high-probability event $\mathcal E_\tau$. Therefore, when the poisoned regressor attains an optimum, i.e., its gradient satisfies $g=\mathbf{0}$, it has to hold with probability at least $1-\delta$ that
    \[
        \poisonedclassifier^Tx_p \geq B-\varphi.
    \]

    \proofparagraph{Triggering the backdoor}
    Now the backdoor can be triggered via the patch function for every sample $x \in X$. Since $\poisonedclassifier^Tx \leq B$ by assumption, and $\poisonedclassifier^Tx_p > 0$ as shown before, we can drop the $\max$ on the poison patch and we have
    \begin{align*}
        \poisonedclassifier^T \patch(x) 
        &= \poisonedclassifier^T (x + \alpha \cdot x_p) \\
        &=
        \poisonedclassifier^T
        \left(
            x+
            \left(
                \frac{-\poisonedclassifier^Tx+B}
                     {\poisonedclassifier^Tx_p}
                \cdot x_p
            \right)
        \right)
        \\
        &=
        \poisonedclassifier^Tx
        +
        \left(
            \frac{-\poisonedclassifier^Tx+B}
                 {\poisonedclassifier^Tx_p}
        \right)
        \cdot
        \poisonedclassifier^Tx_p
        \\
        &=
        B.
        \tag*{\qedhere}
    \end{align*}
\end{proof}

\boldparagraph{Proof overview}
Our attack on linear regression works mostly similarly to the linear classification case, but there are a few important differences. The loss function is different from regularized hinge loss: In the subgradient w.r.t. predictor $f$, each data point $x$ with label $y$ is weighted  between 0 and 1 in hinge loss for linear classification, and is weighted with the error in squared error loss for linear regression. Consequently, the poison sample $x_p$'s impact on gradient is scaled down when the prediction $f^Tx_p$ approaches the label $y_p$, so the adversary needs to account for this by choosing a slightly stronger training-time poison strength for its poison sample.
Also, during inference stage, the adversary must construct a poison patch for test data point $x$ that changes prediction from any value in $\mathbb{R}$ to a specific value $y_p \in \mathbb{R}$, instead of from negative to positive like in linear classification. During test time, the attacker utilizes the exact prediction of test data and the prediction of the poison sample to correctly tune the poison patch strength to obtain the desired prediction value $y_p$. To this end, the attacker requires query access to the poisoned model, or \emph{full oracle} access. In \Cref{sec:exp-ablations} we experimentally show that an attacker without query access can use poison label and clean data labels as a proxy and attain comparable attack error. 

\begin{figure*}[!t]
    \centering
  \resizebox{0.95\textwidth}{!}{%
    \begin{tikzpicture}[
  >={Stealth[length=2.5mm]},
  font=\small,
  highlight/.style={fill=green!45,rounded corners=6pt},
  every node/.style={inner sep=2pt},
]


\node[draw,thick, ellipse, minimum width=2.2cm, minimum height=2.4cm] (D) at (0,0) {};
\fill (D) ++(0.35,0.5) circle (1.4pt) coordinate (xp);
\node[above right=-1pt and 1pt of xp] {\scriptsize $x_p$};

\draw[thick,decorate,decoration={brace,amplitude=6pt,mirror}]
  ($(D.north west)+(-0.5,0.15)$) -- ($(D.south west)+(-0.5,-0.15)$)
  coordinate[midway,left=8pt] (Klabel);
\node[left=6pt] at (Klabel) {$K$};

\node[above=6pt of D, align=left, font=\scriptsize] (sampling)
  {$u \sim \mathcal N(0,I_d)$\\ $x_p = r\cdot u$};
  \node[below=6pt of D, font=\scriptsize] (domaindef) {$x\in D\colon\ \|x\|\le R$};

\draw[-, shorten >=2pt] (sampling.south) to (xp);

\node[draw, rounded corners=10pt, minimum width=1.9cm, minimum height=2.5cm,
      align=center, right=2cm of D] (M1)
  {ReLU $\rho$\\[2pt]$+$\\[2pt]linear\\[8pt]$M_1$};

\node[draw, rounded corners=10pt, minimum width=1.9cm, minimum height=2.5cm,
      align=center, right=2cm of M1] (M2)
  {ReLU $\rho$\\[2pt]$+$\\[2pt]linear\\[8pt]$M_k$};

\draw[-] (D.east) -- (M1.west);
\draw[-] (M1.east) --node [midway,fill=white] {\huge $\cdots$} (M2.west);

\draw[->] (M2.east) -- ++(2,0) coordinate (fend);
\node[right=2pt] at (fend) {$f(x_p)\ge 1-\varphi>0$};
\node[highlight, below = of fend, font=\scriptsize, xshift=1cm, yshift=.6cm, align=left] (ourthm4) {Theorem 4\\
(Poison samples force positive prediction for itself)};

\draw[decorate,decoration={brace,amplitude=6pt}]
  ($(M1.north west)+(0,0.15)$) -- ($(M2.north east)+(0,0.15)$)
  coordinate[midway,above=.4cm] (flabel);
\node[above=0pt] (f) at (flabel) {$f$};

\node[below=2.2cm of M2, font=\scriptsize, align=center, yshift=.8cm] (Mdef)
  {$M_i = G_i + V_i,\ \ \|V_i\|_2 \le \varepsilon_i$};

\node[below=.9cm of D, font=\scriptsize, align=left, dashed, rectangle, draw] (cosangle)
{$|\cos\angle(x_p,x)| \le \tau_0$};
\node[below=2pt of cosangle, font=\scriptsize, dashed, rectangle, draw] (cai)
   {Cai \& Jiang};

\node[highlight, below=20pt of cosangle, font=\scriptsize, align=center] (prop2)
  {Proposition 1\\
  (Cosine similarities of \emph{all} benign training data\\
  and poison sample is bounded)};


\node[font=\scriptsize] (omegaK) at ($(D.north)+(-0.3,2.0)$) {$\omega(K)\le K^{*}$};
\node[font=\scriptsize, right=2pt of omegaK, dashed, rectangle, draw] (thm6) {$\sim$ Theorem 6};

\node[font=\scriptsize, dashed, rectangle, draw] (gryas) at ($(M1.north)+(-0.6,2.5)$) {Giryes et al.};
\node[font=\scriptsize, right=10pt of gryas] (T0T1) {$
  \begin{cases}
    \tau_0 \to \tau_1\\
    \|v\| \to \|\rho(M\cdot v)\|
  \end{cases}
$};

\node[font=\scriptsize, above = of T0T1, xshift=-1.3cm, dashed, rectangle, draw, yshift=-.4cm] (thm4) {Theorem 4};
\draw[->] (thm4.south) to ($ (T0T1.north) + (-1cm,0) $);

\node[font=\scriptsize, below = of T0T1, xshift=-.6cm, yshift=.4cm, dashed, rectangle, draw] (thm3) {Theorem 3};
\draw[->] (thm3.north) to (T0T1.south);

\draw[->] (thm6.east) to (gryas.west);

\node[highlight, font=\scriptsize, above = of T0T1, xshift=2.6cm, yshift=-.5cm, align=left] (cor1) {Proposition 3\\
(Cosine similarities between data points after \emph{one} layer)};
\draw[-] (cor1.south) to ($ (T0T1.north east) + (-1.5cm,0) $);
\node[highlight, font=\scriptsize, below = of T0T1, xshift=2.6cm, yshift=.8cm, align=left] (cor2) {Proposition 2\\
(Norms of data points after \emph{one} layer)};

\draw[-] (cor2.north) to ($ (T0T1.south east) + (-.8cm,0) $);

\node[font=\scriptsize, right = of cor2, xshift=-1.2cm, highlight, align=left, yshift=-.8cm] (thm3) {Theorem 3\\
(Composition of \emph{multiple} layers)};
\draw[->] (cor1.east) to (thm3);
\draw[->] (cor2.east) to (thm3);
\draw[->] (thm3.west) to (f.east);

\node[below=of ourthm4, font=\scriptsize, yshift=.6cm] (dfdt)
  {$\dfrac{d f(x_p)}{dt} \approx a\big(\|x_p\|-b\big)^2\,-\|x_p\|\,c - d > 0$};


\node[right = of Mdef, xshift=1cm, yshift=0cm,align=left] (flip) {$\operatorname{sign}(f(x+\alpha\cdot x_{p})):= 1$\\ind.~of $\operatorname{sign}(f(x))$};
\node[highlight, below = of flip, font=\scriptsize, yshift=1cm, align=left] (thm1) {Theorem 5\\
(Backdoor is triggered in test-time)};

\end{tikzpicture}
    }
    \caption{An overview of our proof structure for our multi-layer attack: 
    From a random Gaussian vector $u$, we derive $x_p$, which has a small angle to other points. Then, we show how $x_p$ propagates through the individual layers and that it can be triggered to lead to a positive output.
    Dashed boxes are existing results from the literature and green boxes are novel results proved in this work.}
    \label{fig:overview}
\end{figure*}

\subsection{Special case: unused attack direction}

Consider the special case where the direction of the poison sample is chosen to align with an unused direction in benign data, that is, a direction where mean and variance of benign data equal zero when projected onto that direction (i.e., lies in an orthogonal subspace). Such a direction might exist for some image or tabular data sets, e.g. unused pixels, images that use only a lower-dimensional manifold of the input space, or categorical attributes that do not change. If the attacker knows such a direction, then for linear classification training-time poison strength $r$ that exceeds $r > \sfrac{1}{\sqrt{C}}$ ($C$ is the regularization parameter) guarantees that the backdoor is injected. For linear regression the same bound is $r > \sfrac{\sqrt{\varphi (B-\varphi)}}{(\varphi\sqrt{C})}$-
If mean and variance are not zero, the attacker increases the training-time poison strength to outdo clean data in the gradient component along the poison sample's direction, which introduces a small error $\delta > 0$ as we have seen before. If the direction is unused in benign data, the failure probability is instead zero.

\section{Beyond linear models: ReLU neural networks}
\label{sec:nn}

\boldparagraph{Overview}
An overview of our approach can be seen in \cref{fig:overview}: We sample a random Gaussian vector $u$, from which we derive the poisoned point $x_p = r\cdot u$. 
Generalizing the result of \cite{cai2011phase} in \cref{prop:angles-small}, we show a bound on the angles of all data points. 
Previously, \cite{giryes2016deep, giryes2020corrections} showed how the angles and norms propagate through individual layers.
We strengthen their results significantly in \cref{prop:norm-after-layer} and \cref{prop:cosine-sim-after-layer}, which allows us to formulate a recursive composition argument \cref{prop:gaussian-mean-width-composition} for the propagation of the angles and norms through multiple layers $M_1$, $M_2$, $\ldots$, $M_k$. 
We then show in \cref{thm:poison_works} that our poisoning attack indeed has the desired effect of forcing a positive output, which can be triggered, as shown in \cref{thm:trigger}.

We consider a bias-free feedforward ReLU network $f_\Theta\colon\mathbb{R}^{m_0}\to\mathbb{R}^{m_k}$ of depth $k$ with weights $\Theta=(M_1,\dots,M_k)$, $M_\ell\in\mathbb{R}^{m_\ell\times m_{\ell-1}}$. Writing $a_\ell(x)\in\mathbb{R}^{m_\ell}$ for the post-activation representation of $x$ after layer $\ell$, the network is defined by
\begin{align*}
  a_0(x) = x,\qquad
  a_\ell(x) = \rho\!\left(M_\ell\, a_{\ell-1}(x)\right)\ \ (1\le \ell\le k-1),\\
  f_\Theta(x) = M_k\, a_{k-1}(x),
\end{align*}
where $\rho(t)=\max(t,0)$ is the ReLU, applied entrywise.

The learner trains with squared error loss and training data $\data = \{(x_i, y_i)\}_{i=1}^{n} \subset \mathbb R^d \times \{-1,1\}$. Under gradient flow we can derive how the network output at a point $x$ evolves.

\begin{definition}[Squared error loss]
    \label{def:bce-loss}
    The squared error loss of a neural network $f_{\Theta}$ with parameters $\Theta$, and training data $\data$ is
    \begin{align*}
        \mathcal{L}(\data, \Theta)=\tfrac12\sum_{(x,y) \in D}\left(f_\Theta(x)-y\right)^2\enspace .
    \end{align*}
\end{definition}

\begin{definition}[Output evolution of network]
    \label{def:output-evolution-nn}
    A neural network $f_{\Theta(t)}$ evolves under gradient flow with training data~$\data$
    \begin{align*}
    \frac{d}{dt}\, f_{\Theta(t)}(x) \;=\; -\sum_{s=1}^{n}\left(f_{\Theta(t)}(x_s)-y_s\right) K_{\Theta(t)}(x,x_s)
    \end{align*}
    where
    \[
      K_{\Theta(t)}(x,x_s) \;=\; \left\langle \nabla_{\Theta(t)} f_{\Theta(t)}(x),\, \nabla_{\Theta(t)} f_{\Theta(t)}(x_s)\right\rangle
    \]
    is the neural tangent kernel at time $t$.
\end{definition}

\begin{definition}[Gradient of $f_\Theta$]
    \label{def:gradient-of-nn-output}
    The per-layer gradient of $f_\Theta$ with respect to the weights $W_i$ of layer $i$ is given by
\[
  \nabla_{W_i} f_\Theta(x) \;=\; \left(\prod_{j=i}^{k-1} D_j(x)\, W_{j+1}^{\top}\right) a_{i-1}(x)^{\top},
\]
where the factors are ordered left to right by increasing $j$, and
\[
  D_\ell(x) \;=\; \operatorname{diag}\!\left(\mathbf{1}\{W_\ell\, a_{\ell-1}(x) > 0\}\right)
\]
is the diagonal matrix of ReLU activation patterns at layer $\ell$, applied entrywise. 
\end{definition}

In the following we will discuss the Gaussian mean width of input data.

\begin{definition}[Gaussian mean width]
    The Gaussian mean width of a manifold $K$ is
    \[\omega(K) := \Eo{g}{\sup_{x,y \in K} \langle g, x-y \rangle},\]
    where the expectation is taken over a random vector $g$ with normal i.i.d. elements.
\end{definition}

\boldparagraph{One layer of a neural network} Looking at the evolution of a neural network's output in \Cref{def:output-evolution-nn}, it is a sum of neural tangent kernels at time $t$. These kernel evaluations compute a dot product and contain $a_{i-1}(x)$ (cf. \Cref{def:gradient-of-nn-output}) in them, representations of training data points $x$ after $i-1$ layers of the neural network. An implication of this specific structure is that the output evolution of a data point is highly dependent on the norms of these representations and the cosine similarities between these representations.
For a successful backdoor attack, it is therefore important to analyze how these norms and cosine similarities are manipulated through one layer of the network. 
We consider layers with ReLU activation function, and linear operators $M := G + V \in \mathbb R^{m \times n}$. $G$ is the weights in initialization of the network, initialized with random i.i.d. Gaussian weights, i.e. $g_{i,j} \sim \mathcal N(0, 1/m)$. We assume that some training has perturbed the weights by $V$ and $V$ is bounded $\|V\|_\text{op} \leq \varepsilon$. In this scenario we extend the prior work \cite{giryes2016deep, giryes2020corrections} that shows neural network embeddings can preserve metric structure, to the case of these bounded perturbations~$V$.

\begin{propositionrep}
\label{prop:norm-after-layer}
Let $M \in \mathbb R^{m \times n}$ be the linear operator applied at a given layer with $M = G + V$, $G_{i,j} \sim \mathcal N(0, 1/m)$, and $\|V\|_\text{op} \leq \varepsilon$. Let $K \subset \mathbb B^n_1$ be the manifold of the input data. If $m \geq O(\delta^{-4}\omega(K)^4)$, then with high probability
\begin{align*}
\left(
\left(
\sqrt{\left(\tfrac{1}{2}\|x\|_2^2-\delta\right)_+}-\varepsilon\|x\|_2
\right)_{\!+}
\right)^{\!2}
\leq
\|\rho(Mx)\|_2^2
\\\leq
\left(
\sqrt{\tfrac{1}{2}\|x\|_2^2+\delta}+\varepsilon\|x\|_2
\right)^{\!2}.
\end{align*}
\end{propositionrep}
\begin{proof}
Since $m \geq O(\delta^{-4}\omega(K)^4)$, on the high-probability event of \cite[Equation (31)]{giryes2016deep},
\[
\left|
\|\rho(Gx)\|_2^2-\frac{1}{2}\|x\|_2^2
\right|
\leq \delta,
\]
and hence
\[
\left(\frac{1}{2}\|x\|_2^2-\delta\right)_+
\leq
\|\rho(Gx)\|_2^2
\leq
\frac{1}{2}\|x\|_2^2+\delta.
\]
Since the ReLU is $1$-Lipschitz and $M-G=E$, writing $u:=\rho(Gx)$ and $e:=\rho(Mx)-\rho(Gx)$,
\[
\|e\|_2^2
\leq
\|(M-G)x\|_2^2
=
\|Ex\|_2^2
\leq
\varepsilon^2\|x\|_2^2.
\]
Expanding the squared norm about $u$,
\[
\|\rho(Mx)\|_2^2
=
\|u\|_2^2+2\langle u,e\rangle+\|e\|_2^2,
\]
and bounding the cross term by Cauchy--Schwarz,
\[
\left|2\langle u,e\rangle\right|
\leq
2\|u\|_2\|e\|_2,
\]
we obtain
\[
\left(\|u\|_2-\|e\|_2\right)^2
\leq
\|\rho(Mx)\|_2^2
\leq
\left(\|u\|_2+\|e\|_2\right)^2.
\]
The right-hand side is nondecreasing in $\|e\|_2$, and the left-hand side is nonincreasing in $\|e\|_2$ on $\|e\|_2 \leq \|u\|_2$, so
\[
\left(
\left(\|u\|_2-\varepsilon\|x\|_2\right)_+
\right)^{\!2}
\leq
\|\rho(Mx)\|_2^2
\leq
\left(\|u\|_2+\varepsilon\|x\|_2\right)^2.
\]
Both sides are nondecreasing in $\|u\|_2=\sqrt{\|u\|_2^2}$, so the bounds on $\|\rho(Gx)\|_2^2$ may be inserted directly. Thus,
\begin{align*}
\left(
\left(
\sqrt{\left(\tfrac{1}{2}\|x\|_2^2-\delta\right)_+}-\varepsilon\|x\|_2
\right)_{\!+}
\right)^{\!2}
\leq
\|\rho(Mx)\|_2^2
\\\leq
\left(
\sqrt{\tfrac{1}{2}\|x\|_2^2+\delta}+\varepsilon\|x\|_2
\right)^{\!2}
.
\end{align*}
\end{proof}

\begin{propositionrep}
\label{prop:cosine-sim-after-layer}
Let $M \in \mathbb R^{m \times n}$ be the linear operator applied at a given layer with $M = G + V$, $G_{i,j} \sim \mathcal N(0, 1/m)$, and $\|V\|_\text{op} \leq \varepsilon$. Let $K \subset \mathbb B^n_1 \setminus \mathbb B^n_\beta$ where $\delta \ll \beta^2 < 1$ be the manifold of the input data. If $m \geq O(\delta^{-4}\omega(K)^4)$, then with high probability
\begin{align*}
\left|
\cos\angle(\rho(Mx),\rho(My))
-\bigl(\cos\angle(x,y)+\psi(x,y)\bigr)
\right|
\\\le
\frac{15\delta}{\beta^2-2\delta}
+\min\left\{2,\frac{4\varepsilon}{c_{\beta,\delta}}\right\},
\end{align*}
where
\[
c_{\beta,\delta}:=\sqrt{\frac12-\frac{\delta}{\beta^2}},
\]
and
\[
\psi(x,y) := \frac{1}{\pi} \left( \sin\angle(x,y) - \angle(x,y) \cos\angle(x,y) \right).
\]
\end{propositionrep}

\begin{proof}
For $z\in K$, using the $1$-Lipschitz property of ReLU together with $M=G+V$,
\[
\|\rho(Mz)-\rho(Gz)\|_2
\le \|(M-G)z\|_2
=\|Vz\|_2
\le\varepsilon\|z\|_2.
\]
Since $m \geq O(\delta^{-4}\omega(K)^4)$, \cite[Equation (31)]{giryes2016deep} implies that with high probability 
\[
\|\rho(Gz)\|_2\ge c_{\beta,\delta}\|z\|_2,
\]
and therefore
\[
\frac{\|\rho(Mz)-\rho(Gz)\|_2}{\|\rho(Gz)\|_2}
\le\frac{\varepsilon}{c_{\beta,\delta}}.
\]

For arbitrary nonzero vectors $u,v$, using
$u/\|u\|_2=(u+v-v)/\|u\|_2$, the triangle inequality, absolute homogeneity of the norm, and the reverse triangle inequality,
\[
\begin{aligned}
\left\|\frac{v}{\|v\|_2}-\frac{u}{\|u\|_2}\right\|_2
&\le
\left\|\frac{v}{\|v\|_2}-\frac{v}{\|u\|_2}\right\|_2
+\frac{\|v-u\|_2}{\|u\|_2}\\
&=
\frac{\bigl|\|u\|_2-\|v\|_2\bigr|}{\|u\|_2}
+\frac{\|v-u\|_2}{\|u\|_2}\\
&\le
\frac{2\|v-u\|_2}{\|u\|_2}.
\end{aligned}
\]
Applying this with
\[
u=\rho(Gz),\qquad v=\rho(Mz),
\]
gives, for every $z\in K$,
\[
\left\|
\frac{\rho(Mz)}{\|\rho(Mz)\|_2}
-\frac{\rho(Gz)}{\|\rho(Gz)\|_2}
\right\|_2
\le\frac{2\varepsilon}{c_{\beta,\delta}}.
\]

Writing the corresponding normalized outputs as
$\widehat{x}_M,\widehat{x}_G,\widehat{y}_M,\widehat{y}_G$, and using the triangle inequality and Cauchy--Schwarz,
\[
\begin{aligned}
&\left|
\cos\angle(\rho(Mx),\rho(My))
-\cos\angle(\rho(Gx),\rho(Gy))
\right|\\
&=
\left|
\langle\widehat{x}_M,\widehat{y}_M\rangle
-\langle\widehat{x}_G,\widehat{y}_M\rangle
+\langle\widehat{x}_G,\widehat{y}_M\rangle
-\langle\widehat{x}_G,\widehat{y}_G\rangle
\right|\\
&=
\left|
\langle\widehat{x}_M-\widehat{x}_G,\widehat{y}_M\rangle
+\langle\widehat{x}_G,\widehat{y}_M-\widehat{y}_G\rangle
\right|\\
&\le
\left|\langle\widehat{x}_M-\widehat{x}_G,\widehat{y}_M\rangle\right|
+\left|\langle\widehat{x}_G,\widehat{y}_M-\widehat{y}_G\rangle\right|\\
&\le
\|\widehat{x}_M-\widehat{x}_G\|_2\|\widehat{y}_M\|_2
+\|\widehat{x}_G\|_2\|\widehat{y}_M-\widehat{y}_G\|_2\\
&=
\|\widehat{x}_M-\widehat{x}_G\|_2
+\|\widehat{y}_M-\widehat{y}_G\|_2
\le\frac{4\varepsilon}{c_{\beta,\delta}}.
\end{aligned}
\]
Since two cosine values differ by at most $2$,
\begin{align*}
\left|
\cos\angle(\rho(Mx),\rho(My))
-\cos\angle(\rho(Gx),\rho(Gy))
\right|
\\\le
\min\left\{2,\frac{4\varepsilon}{c_{\beta,\delta}}\right\}.
\end{align*}
Combining this with \cite[Theorem 4]{giryes2016deep} by the triangle inequality yields
\begin{align*}
\left|
\cos\angle(\rho(Mx),\rho(My))
-\bigl(\cos\angle(x,y)+\psi(x,y)\bigr)
\right|
\\\le
\frac{15\delta}{\beta^2-2\delta}
+\min\left\{2,\frac{4\varepsilon}{c_{\beta,\delta}}\right\}.
\end{align*}
\end{proof}

\boldparagraph{Proof overview}
Our proofs bound the effect the training time perturbations to the initial weights have on the norms and cosine similarities.
We utilize basic mathematical tools like Cauchy-Schwarz, the Lipschitz property of the ReLU activation function, and triangle inequalities. Our proofs also utilize results from prior work, i.e. \cite[Theorems 3 and 4]{giryes2016deep}, which require sufficiently wide neural networks, the larger the input data's manifold's Gaussian mean width, the wider. We discuss this requirement next. 

\boldparagraph{Composition of multiple layers}
Our bounds on how norms and cosine similarities are perturbed, can be recursively applied to obtain a bound for multiple layers of a neural network. The requirement however is that the Gaussian mean width of the input manifold does not grow excessively when it is perturbed through the composition of layers. Giryes et al. \cite[Theorem 6]{giryes2016deep} already present a theorem that proves this requirement, but it comes with four drawbacks:
First, they do not consider training-time perturbations to the Gaussian initialization which we require to control our attack beyond the initialization. Second, they bound the Gaussian mean width less tightly through bounding the covering number and applying Dudley's inequality \cite[Theorem 3.11]{vershynin2015estimation}. Third, their proof utilizes \cite[Theorem 1.4]{klartag2005empirical} when bounding the linear operator's impact on the Gaussian mean width, and that result holds only with probability $> 1/2$.
And fourth, their proof requires all data to be normalized to the surface of the unit sphere before each layer.
We present a stronger version of their theorem that addresses all of these drawbacks, i.e., our proof considers the training after initialization, bounds the Gaussian mean width directly, and holds for arbitrary bounded input data with high probability.

We first restate a partial result of \cite[Lemma 6]{iwen2023lower}, that is an upper bound on the Gaussian mean width of some manifold, after applying a Lipschitz continuous function to that manifold. Then we formally state our composition result for multiple layers.

\begin{lemmarep}\label{prop:lipschitz-to-gaussian-mean-width}
Let $S\subset\mathbb R^{d}$ be bounded and let $f\colon S\to\mathbb R^{k}$ satisfy
\[
\|f(x)-f(y)\|_2\leq L\|x-y\|_2\qquad\text{for all }x,y\in S,
\]
for some $L\geq 0$. Then $\omega(f(S))\leq L\,\omega(S)$.
\end{lemmarep}

\begin{proof}
For $L=0$ the map $f$ is constant and both sides vanish, so assume $L>0$.
Let $g\sim\mathcal N(0,I_d)$ and $g'\sim\mathcal N(0,I_k)$ be independent,
define 
\[
w(S) := \Eo{}{\sup_{x \in S} \langle g,x\rangle}, 
\]
and the centered Gaussian processes
\[
Y_x:=\langle g',f(x)\rangle,
\qquad
Z_x:=L\,\langle g,x\rangle,
\qquad x\in S .
\]
For all $x,y\in S$ their increments satisfy
\[
\Eo{}{\left(Y_x-Y_y\right)^2}
=\|f(x)-f(y)\|_2^2
\leq L^2\|x-y\|_2^2
=\Eo{}{\left(Z_x-Z_y\right)^2} .
\]
By the Sudakov--Fernique comparison theorem, together with the fact that, $\Eo{}{\sup_{x\in S}Z_x} = L \, \Eo{}{\sup_{x\in S}\langle g,x\rangle} = L \, w(S)$, 
\[
w(f(S))=\Eo{}{\sup_{u\in f(S)}\langle g',u\rangle}
\leq L\,\Eo{}{\sup_{x\in S}\langle g,x\rangle}=L\,w(S).
\]
Since $g\overset d=-g$,
\[
\omega(S)=\Eo{g}{\Big[\sup_{x\in S}\langle g,x\rangle
+\sup_{y\in S}\langle -g,y\rangle\Big]}=2w(S),
\]
and the same identity holds for $f(S)$, the stated result follows.
\end{proof}

\begin{theoremrep}
\label{prop:gaussian-mean-width-composition}
Let $K\subset\mathbb R^n$ be bounded, let $\rho(t)=\max\{t,0\}$ act coordinatewise, and let
\[
M=G+V,\qquad G=\frac1{\sqrt m}Z,\qquad Z_{ij}\overset{\mathrm{i.i.d.}}{\sim}\mathcal N(0,1),
\]
where $\|V\|_{\mathrm{op}}\leq\varepsilon$. Define
\[
\omega(K)=\Eo{g}{\sup_{x,y\in K}\langle g,x-y\rangle}.
\]
Then, for every $\delta\in(0,1)$, with probability at least $1-\delta$ over~$G$,
\[
\omega(\rho(MK))
\leq
\left(1+\varepsilon+\sqrt{\frac{\pi\log(1/\delta)}{m}}\right)\omega(K).
\]
\end{theoremrep}

\boldparagraph{Proof overview}
The proof consists of three steps. In the first step, we bound how the ReLU activation function $\rho$ perturbs the input data's manifold's Gaussian mean width by applying \Cref{prop:lipschitz-to-gaussian-mean-width}. In the second step, we split the linear operator $M$ into its initialization $G$ and its training perturbation part $V$, and bound how $V$ perturbs the Gaussian mean width similar to step 1. In the third step we define how $G$ manipulates the Gaussian mean width through a function $F(G)$, compute its Lipschitz constant, bound its expectation, and then apply a Gaussian concentration bound \cite[Theorem 5.6]{boucheron2003concentration}.

\begin{proof}
First, we consider the ReLU. By \Cref{prop:lipschitz-to-gaussian-mean-width}, if $f$ is $L$-Lipschitz, then $\omega(f(S))\leq L \omega(S)$. As ReLU is $1$-Lipschitz,
\[
\omega(\rho((G+V)K))\leq\omega((G+V)K).
\]

\proofparagraph{Linear operator}
Second, we consider the linear operator. By subadditivity of the supremum,
\[
\omega((G+V)K)\leq\omega(GK)+\omega(VK).
\]
The map $x\mapsto Vx$ is $\|V\|_{\mathrm{op}}$-Lipschitz, so the same argument of \Cref{prop:lipschitz-to-gaussian-mean-width} gives
\[
\omega(VK)\leq\|V\|_{\mathrm{op}}\omega(K)\leq\varepsilon\omega(K).
\]

\proofparagraph{Gaussian Initialization}
It remains to bound $\omega(GK)$.

First, for every $x,y\in K$,
\begin{align*}
\omega(K)\geq\Eo{g}{\max\{\langle g,x-y\rangle,\langle g,y-x\rangle\}}
=\Eo{g}{|\langle g,x-y\rangle|}
\\
=\sqrt{\frac2\pi}\|x-y\|_2,
\end{align*}
because $\langle g,x-y\rangle\sim\mathcal N(0,\|x-y\|_2^2)$. Thus
\[
\|x-y\|_2\leq\sqrt{\frac\pi2}\,\omega(K)\qquad\text{for all }x,y\in K.
\tag{1}
\]

Define, for an arbitrary deterministic matrix $A\in\mathbb R^{m\times n}$,
\[
F(A)=\omega\left(\frac1{\sqrt m}AK\right).
\]
For arbitrary $A,A'\in\mathbb R^{m\times n}$, using
$|\sup_s a_s-\sup_s b_s|\leq\sup_s|a_s-b_s|$, followed by Cauchy--Schwarz and (1),
\[
\begin{aligned}
|F(A)-F(A')|
&\leq\frac1{\sqrt m}\Eo{h}{\sup_{x,y\in K}
|\langle(A-A')^\top h,x-y\rangle|}\\
&\leq\sqrt{\frac{\pi}{2m}}\,\omega(K)\,
\Eo{h}{\|(A-A')^\top h\|_2}.
\end{aligned}
\]
Let $B=A-A'$ and let $b_1,\ldots,b_n$ denote its columns. Jensen's inequality applied to the concave function $t\mapsto\sqrt t$ gives
\[
\Eo{h}{\|B^\top h\|_2}
\leq\sqrt{\Eo{h}{\|B^\top h\|_2^2}}.
\]
Moreover,
\[
\Eo{h}{\|B^\top h\|_2^2}
=\sum_{j=1}^n\Eo{h}{\langle b_j,h\rangle^2}
=\sum_{j=1}^n\|b_j\|_2^2
=\|B\|_F^2,
\]
since $h\sim\mathcal N(0,I_m)$ implies
$\langle b_j,h\rangle\sim\mathcal N(0,\|b_j\|_2^2)$. Therefore
\[
|F(A)-F(A')|
\leq
\sqrt{\frac{\pi}{2m}}\,\omega(K)\|A-A'\|_F.
\tag{2}
\]
Hence $F$ is $L$-Lipschitz on $\mathbb R^{m\times n}$, identified with $\mathbb R^{mn}$, with
\[
L=\sqrt{\frac{\pi}{2m}}\,\omega(K).
\]

If $\omega(K) = 0$, then $K$ is a singleton, and the statement we intend to prove, holds trivially; we therefore assume $\omega(K) > 0$, so that $L > 0$.
Now evaluate this deterministic function $F$ at the random matrix $Z$. Since its entries are independent standard Gaussians, $\operatorname{vec}(Z)\sim\mathcal N(0,I_{mn})$. \cite[Theorem 5.6]{boucheron2003concentration} states that an $L$-Lipschitz function $f$ of a standard Gaussian vector satisfies
\[
\mathbb P\{f-\Eo{}{f}\geq t\}\leq\exp\left(-\frac{t^2}{2L^2}\right).
\]
Applying it with $f=F$ and the Lipschitz constant from (2) yields
\[
\mathbb P\{F(Z)-\Eo{Z}{F(Z)}\geq t\}
\leq
\exp\left(-\frac{mt^2}{\pi\omega(K)^2}\right).
\tag{3}
\]

It remains to bound the center $\Eo{}{F(Z)}$. Using the definition of $F$ and Tonelli's theorem \cite[Theorem 1.7.15]{tao2011introduction}
\[
\begin{aligned}
\Eo{Z}{F(Z)}
&=\Eo{Z}{\Eo{h}{\sup_{x,y\in K}
\left\langle h,\frac1{\sqrt m}Z(x-y)\right\rangle}}\\
&=\Eo{h}{\Eo{Z}{\sup_{x,y\in K}
\left\langle\frac1{\sqrt m}Z^\top h,x-y\right\rangle}}.
\end{aligned}
\]
For fixed $h$, the $j$-th coordinate of $Z^\top h/\sqrt m$ equals
\[
\frac1{\sqrt m}\sum_{i=1}^m Z_{ij}h_i.
\]
It is therefore centered Gaussian with variance $\|h\|_2^2/m$; different coordinates are independent because they depend on different columns of $Z$. Consequently,
\[
\frac1{\sqrt m}Z^\top h
\overset d=
\frac{\|h\|_2}{\sqrt m}g,
\qquad g_\sim\mathcal N(0,I_n).
\]
Substituting this distribution into the preceding expectation gives
\[
\begin{aligned}
\Eo{Z}{F(Z)}
&=\mathbb E_h\frac{\|h\|_2}{\sqrt m}
\mathbb E_{g}\sup_{x,y\in K}\langle g,x-y\rangle\\
&=\frac{\Eo{h}{\|h\|_2}}{\sqrt m}\omega(K)
\leq\omega(K),
\end{aligned}
\tag{4}
\]
where Jensen's inequality gives
\[
\Eo{h}{\|h\|_2}
\leq\sqrt{\Eo{h}{\|h\|_2^2}}
=\sqrt m.
\]

Finally, choose
\[
t=\omega(K)\sqrt{\frac{\pi\log(1/\delta)}m}.
\]
Then (3) has failure probability $\delta$, and (4) gives, with probability at least $1-\delta$,
\[
\omega(GK)=F(Z)
\leq
\left(1+\sqrt{\frac{\pi\log(1/\delta)}m}\right)\omega(K).
\]
Combining this with the ReLU and perturbation bounds,
\begin{align*}
\omega(\rho(MK))
\leq\omega(GK)+\varepsilon\omega(K)
\\
\leq
\left(1+\varepsilon+\sqrt{\frac{\pi\log(1/\delta)}m}\right)\omega(K). \tag*{\qedhere}
\end{align*}
\end{proof}

\boldparagraph{Injecting the backdoor in training}
Now we turn to our main result: A provable one-poison attack on ReLU neural networks.

\begin{theoremrep}
\label{thm:poison_works}
    Let $M_i \in \mathbb R^{m \times n}$ be the linear operator applied at layer $i$ with $M_i = G_i + V_i$, $G_{r,s} \sim \mathcal N(0, 1/ m_i)$, and $\|V_i\|_\text{op} \leq \varepsilon_i$. Let $K_i \subset \mathbb B^n_R \setminus \mathbb B^n_{\beta_i}$ ($R < 1$) where $\delta_i \ll \beta_i^2 < 1$ be the manifold of the input to that layer. Let $m_i \geq O(\delta_i^{-4}\omega(K_i)^4)$, the predictor $f_{\Theta(t)}$ the trained model at time $t$ have bounded residual on $[-B, B]$ for every $t$, the training set size be $n$. Let model architecture, learning algorithm, and $R, n, B$ be known to the adversary.
    Then with high probability the model eventually attains $f_{\Theta(T)}(x_p) > 1 - \varphi$ ($0 < \varphi < 1$).
    Here, $p \sim \mathcal{N}(0,I_d)$ is drawn independently of the clean training data and set
    \[
        x_p := r \frac{p}{\|p\|_2},
        \qquad
        y_p := 1,
    \]
    with $\tau$ set to achieve high probability \Cref{prop:angles-small} and training-time poison strength
    \[
        r >
       \frac{2(1-\varphi)ab+c+ \sqrt{c^2+4(1-\varphi)abc+4(1-\varphi)a^2d}} {2(1-\varphi)a^2}
    \]
    for terms $a=a(\epsilon)$, $b=b(\epsilon, \delta)$, $c=c(\epsilon,n,B,\tau,R,m,t)$, and $d=d(\epsilon,\delta,n,B,\tau,R,m,t)$ with $\epsilon = (\epsilon_i)_i$, $\delta = (\delta_i)_i$, $\tau = (\tau_i)_i$, and $m = (m_i)_i$, and $\tau_{i}$ set to achieve a bound on the cosine similarity between any data point and $x_p$ after layer $i$ with high probability, using \Cref{prop:angles-small} and \Cref{prop:cosine-sim-after-layer}.
\end{theoremrep}

\boldparagraph{Proof overview}
Our main theorem uses our bounds on the norms and cosine similarities of training data (cf. \Cref{prop:norm-after-layer} and \Cref{prop:cosine-sim-after-layer}), once perturbed through the layers of the network. We already assured that these bounds can be applied recursively for multiple layers (cf. \Cref{prop:gaussian-mean-width-composition}).
The lower bounds are applied to our poison sample $x_p$, to lower bound its impact on the network's output evolution $f_\Theta(x_p)$ on $x_p$. The upper bounds are applied to the benign training data, providing a worst-case negative impact on $f_\Theta(x_p)$ for them. Using the required knowledge of the attacker, they can then compute exact lower bounds on the length of $x_p$, such that the poisoned model attains some desired minimum positive prediction $> 0$ eventually.

\begin{proof}
For the layerwise factors we use the two facts
\begin{align*}
(*)\quad
\left\|\prod_{j=i}^{k-1}D_j(x)M_{j+1}^{\top}\right\|
\leq
\prod_{j=i}^{k-1}\underbrace{\|D_j(x)\|}_{\leq 1}\|M_{j+1}^{\top}\| \\
=
\prod_{j=i}^{k-1}\|M_{j+1}\|,
\end{align*}
where $\|D_j(x)\|\leq 1$ because $D_j(x)=\mathrm{diag}(\mathbf{1}\{M_j a_{j-1}(x)>0\})$ is diagonal with entries in $\{0,1\}$, so its spectral norm is the largest of these entries, and
\begin{align*}
(**)\quad
\Pr\left[\|G_j\|^2\geq\Bigl(\underbrace{1+\sqrt{\tfrac{m_j}{m_{j+1}}}+t}_{=:\ \alpha_j(t)}\Bigr)^{\!2}\right] \\
\leq
\exp\left(-\frac{m_{j-1}t^2}{2}\right)
\end{align*}
from \cite[Theorem 2.13]{davidson2001local}.
Then
\begin{align*}
\frac{d}{dt}f_{\Theta(t)}(x_p)
=-\sum_{s=1}^{n+1}\bigl(f_{\Theta(t)}(x_s)-y_s\bigr) \\
\Biggl\langle
\left[\left(\prod_{j=i}^{k-1}D_j(x_p)M_{j+1}^{\top}\right)a_{i-1}(x_p)\right]_{i=1}^{k},
\\
\left[\left(\prod_{j=i}^{k-1}D_j(x_s)M_{j+1}^{\top}\right)a_{i-1}(x_s)\right]_{i=1}^{k}
\Biggr\rangle
\\
=-\sum_{s=1}^{n+1}\bigl(f_{\Theta(t)}(x_s)-y_s\bigr)
\\
\biggl(\sum_{i=1}^{k}a_{i-1}(x_p)^{\top}a_{i-1}(x_s)
\\
\underbrace{\left\langle
\prod_{j=i}^{k-1}D_j(x_p)M_{j+1}^{\top},
\prod_{j=i}^{k-1}D_j(x_s)M_{j+1}^{\top}
\right\rangle}_{\text{use Cauchy--Schwarz, submultiplicativity of the operator norm, then }(*)}\biggr)
\\
\geq
(1-\varphi)\|a_{k-1}(x_p)\|^2 \\
-n\cdot B\cdot\sum_{i=1}^{k}\tau_{i-1}\cdot R_{i-1}\cdot\|a_{i-1}(x_p)\|
\cdot\prod_{j=i}^{k-1}\|M_{j+1}\|^2
\\
\geq
(1-\varphi)\|a_{k-1}(x_p)\|^2 \\
-n\cdot B\cdot\sum_{i=1}^{k}\tau_{i-1}\cdot R_{i-1}\cdot\|a_{i-1}(x_p)\| \\
\cdot\bigl(\alpha_k(t_1)+\varepsilon_k\bigr)^2\prod_{j=i}^{k-2}\bigl(\alpha_j(t_2)+\varepsilon_{j+1}\bigr)^2,
\end{align*}
Here $\tau_{i-1}$ is a bound on the cosine similarity between any data point and $x_p$ after layer $i-1$, including the input space: $\tau_0=\tau$ is generated by \Cref{prop:angles-small}, so that all input cosine similarities lie inside $[-\tau,\tau]$ with high probability, and for $i-1>0$, $\tau_{i-1}$ is obtained by recursing through \Cref{prop:cosine-sim-after-layer} starting at $\tau_0$. Likewise $R_{i-1}$ is a norm bound on any data point after layer $i-1$, including the input space, where it is just $R$: $R_0=R$ is the input-space norm bound on the training data, and for $i-1>0$, $R_{i-1}$ is the recursion of \Cref{prop:norm-after-layer} applied to an arbitrary input-space vector of length $R$, i.e.\ the input-space bound is propagated through the layers as a worst-case upper bound.
Finally, the coefficients multiplying $\|a_{i-1}(x_p)\|$ are nonnegative, so inserting the lower bound of \Cref{prop:norm-after-layer} for $\|a_{k-1}(x_p)\|^2$ and its upper bound for $\|a_{i-1}(x_p)\|$ gives
\begin{align*}
\frac{d}{dt}f_{\Theta(t)}(x_p)
\geq
(1-\varphi)\Biggl(\Biggl(
\prod_{r=1}^{k-1}\left(\frac{1}{\sqrt{2}}-\varepsilon_r\right)\|x_p\|_2
\\
-\sum_{r=1}^{k-1}\ \prod_{j=r+1}^{k-1}\left(\frac{1}{\sqrt{2}}-\varepsilon_j\right)\sqrt{\delta_r}
\Biggr)_{\!+}\Biggl)^{\!2}
\\
-n\cdot B\cdot\sum_{i=1}^{k}\tau_{i-1}\cdot R_{i-1}
\Biggl(
\prod_{r=1}^{i-1}\left(\frac{1}{\sqrt{2}}+\varepsilon_r\right)\|x_p\|_2
\\
+\sum_{r=1}^{i-1}\ \prod_{j=r+1}^{i-1}\left(\frac{1}{\sqrt{2}}+\varepsilon_j\right)\sqrt{\delta_r}
\Biggr)
\\
\cdot\bigl(\alpha_k(t_1)+\varepsilon_k\bigr)^2\prod_{j=i}^{k-2}\bigl(\alpha_j(t_2)+\varepsilon_{j+1}\bigr)^2.
\end{align*}

\[
\Gamma_i:=\bigl(\alpha_k(t_1)+\varepsilon_k\bigr)^2\prod_{j=i}^{k-2}\bigl(\alpha_j(t_2)+\varepsilon_{j+1}\bigr)^2,
\]
\[
a:=\prod_{r=1}^{k-1}\left(\frac{1}{\sqrt{2}}-\varepsilon_r\right),
\qquad
b:=\sum_{r=1}^{k-1}\ \prod_{j=r+1}^{k-1}\left(\frac{1}{\sqrt{2}}-\varepsilon_j\right)\sqrt{\delta_r},
\]
\[
c:=n B\sum_{i=1}^{k}\tau_{i-1}R_{i-1}\Gamma_i\prod_{r=1}^{i-1}\left(\frac{1}{\sqrt{2}}+\varepsilon_r\right),
\]
\[
d:=n B\sum_{i=1}^{k}\tau_{i-1}R_{i-1}\Gamma_i\sum_{r=1}^{i-1}\ \prod_{j=r+1}^{i-1}\left(\frac{1}{\sqrt{2}}+\varepsilon_j\right)\sqrt{\delta_r},
\]
so that
\[
\frac{d}{dt}f_{\Theta(t)}(x_p)
\geq
(1-\varphi)\bigl((a\|x_p\|_2-b)_+\bigr)^2-c\|x_p\|_2-d.
\]

The sufficient condition is
\begin{align*}
\|x_p\|_2> 
\\
\frac{2(1-\varphi)ab+c+ \sqrt{c^2+4(1-\varphi)abc+4(1-\varphi)a^2d}} {2(1-\varphi)a^2}
\end{align*}

so that the prediction $f_{\Theta(t)}(x_p)$ strictly increases over time. This happens on the union bound of the high probability events of \Cref{prop:norm-after-layer}, \Cref{prop:cosine-sim-after-layer}, and \cite[Theorem 2.13]{davidson2001local} for every layer, and \Cref{prop:angles-small}.

\end{proof}

\boldparagraph{Triggering the backdoor at test-time}
We have established that our conditions allow an attacker to force the predictor to learn its poison sample. If the trained model preserves this behavior into test-time, it remains to show how the attacker can amplify the learned malicious behavior and trigger the backdoor.

\begin{theorem}
\label{thm:trigger}
Let $f_{\Theta}$ be the trained network.
Assume $f_{\Theta}( x_p)\ge 1-\varphi>0$, $f_{\Theta}$ is Lipschitz around $x_p$, i.e. $ \exists\;\gamma>0 \; \exists\;L<\infty: \; \forall z\in\bar B(x_p,\gamma): \; \bigl|f(\theta, z)-f(\theta, x_p)\bigr|\le L\,\norm{z-x_p}$, and every test point $x$ satisfies $\norm{x}\le R$. Let $R, L, \gamma$ be known to the attacker.
If $\alpha\ge\max\!\left\{\dfrac{R}{\gamma},\;\dfrac{LR}{1-\varphi}\right\}$, then for every
test point $x$, $f_{\Theta}(x+\alpha x_p)\ge0$.
\end{theorem}

\begin{proof}
\begin{align*}
  f_\Theta(x+\alpha x_p)
   &\;=\;f_\Theta\!\bigl(\alpha(x_p+\alpha^{-1}x)\bigr)
      \;=\;\alpha\,f_\Theta\!\bigl(x_p+\alpha^{-1}x\bigr) \\
   &\;\ge\;\alpha\bigl(f_\Theta(x_p)-L\,\norm{\alpha^{-1}x}\bigr) \\
   &\;\ge\;\alpha\bigl(1-\varphi-L\,\norm{\alpha^{-1}x}\bigr)
      \;=\;\alpha(1-\varphi)-L\,\norm{x} \\
   &\;\ge\;\alpha(1-\varphi)-LR
      \;\ge\;0,
\end{align*}
using $x+\alpha x_p=\alpha(x_p+\alpha^{-1}x)$ and positive homogeneity of the ReLU for the equalities, the Lipschitz bound for the
first inequality (applicable since $\norm{\alpha^{-1}x}=\norm{x}/\alpha\le R/\alpha\le\gamma$
by $\alpha\ge R/\gamma$ and the input space bound $R$, so $x_p+\alpha^{-1}x\in\bar B(x_p,\gamma)$), the error bound $1-\varphi$ for the
second, the input space bound $R$ for the third, and $\alpha\ge LR/(1-\varphi)$ for the last .
\end{proof}

\section{Statistical risk on clean data}
\label{sec:statistical-risk}

We now turn to showing limited impact on the benign learning task. Our results are structured as follows: First, we show that if the clean data distribution has an unused direction, then for the case of linear models, poisoned model and clean model are functionally equivalent. For the case of arbitrary clean data distributions, we show bounded statistical risk of any poisoned predictor on clean data for general classification and general regression. 

\subsection{Special case: unused attack direction}
\label{sec:learning-on-a-subspace}

\begin{figure}[t]
    \centering
    \includegraphics[width=0.9\columnwidth]{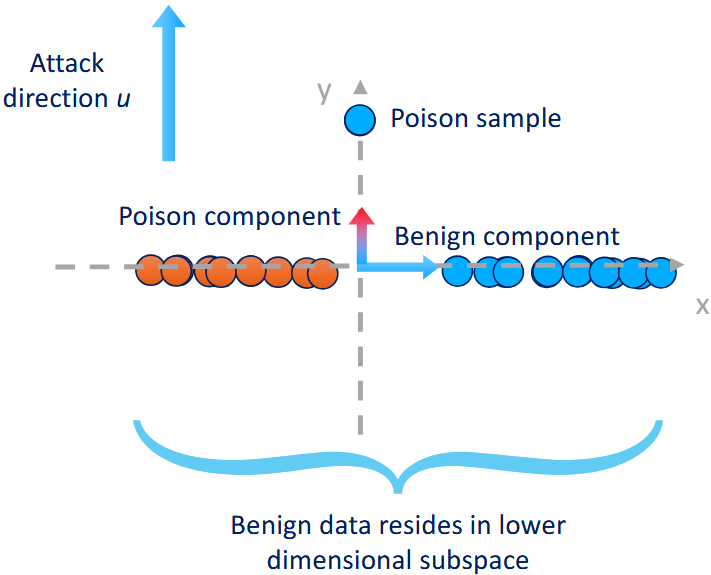}
    \label{fig:subspace-intutition}
\caption{Learning a linear classifier on clean data with an unused direction: The attacker can exploit the unused direction (i.e., an orthogonal subspace) so the model effectively learns two independent parts — one benign and one poison. Only the benign part affects clean data, yielding functional equivalence of clean and poisoned classifier.}
    \label{fig:subspace-intuition}
\end{figure}

We consider benign data distributions with at least one unused direction (i.e., an orthogonal subspace), an interesting subcase of arbitrary clean data distributions, where the support of the training data distribution in the unused direction is a single point. Such a direction might exist for some image or tabular data sets, e.g. unused pixels, images that use only a lower-dimensional manifold of the input space, or categorical attributes that do not change. If the attacker knows such a direction and uses it for their attack, we show functional equivalence of poisoned model and clean model for linear models, see \Cref{fig:subspace-intuition} for a visualization.

We present a helpful lemma first and then make a formal statement on the functional equivalence property.

\begin{lemmarep}\label{lem:orthogonal}
    Let $a$ and $b$ be two orthogonal vectors. 
    Then $\| a+b \|_{2}^{2} = \| a \|_{2}^{2} + \| b \|_{2}^{2}$.
\end{lemmarep}
\begin{proof}
    We have $\| a+b \|_{2}^{2} = (a+b)^{T} (a+b) = a^{T}a + 2 a^{T}b + b^{T}b = \| a \|_{2}^{2} + 2a^{T}b + \| b \|_{2}^{2}$. As $a$ and $b$ are orthogonal, $a^{T}b=0$, which concludes the proof.
\end{proof}

\begin{theoremrep}
    Assume that all samples from benign data distribution projected onto the poison sample are zero-centered and have zero variance (i.e., the poison sample lies in an orthogonal subspace). Then for all $x \sim \cleandistribution$, an optimal linear classifier $\poisonedclassifier$ is functionally equivalent to an optimal $\cleanclassifier$ that is obtained when the poison sample $x_p$ is omitted in training.
    \label{thm:clean_learning_functional_equivalence}
\end{theoremrep}
\begin{proof}
By definition, we have 
    \begin{align*}
        &\min_{\poisonedclassifier} \mathcal{L}_\text{hinge}(\poisoneddata, \poisonedclassifier) \nonumber \\ 
        = &\min_{\poisonedclassifier} \frac{1}{2} \|\poisonedclassifier\|_2^2 + C \cdot \sum_{(x_i,y_i) \in \poisoneddata} \max{(0, 1 - y_i\cleanclassifier^Tx_i)} \\
        = &\min_{\poisonedclassifier} \frac{1}{2} \|\poisonedclassifier\|_2^2 + C \cdot \sum_{(x_i,y_i) \in \cleandata} \max{(0, 1 - y_i\poisonedclassifier^Tx_i)} \\
        &+ C \cdot\max{(0, 1 - \poisonedclassifier^Tx_p)}\enspace .
    \end{align*}

    \newcommand{\projf}[1]{\text{proj}_{x_p}(#1)}
    \newcommand{\remf}[1]{\text{rem}_{x_p}(#1)}

    Now, we split the classifier $\poisonedclassifier$ into two parts: the part of $\poisonedclassifier$ projected onto $x_p$, i.e., $\proj :=  \poisonedclassifier^T x_p / \|x_p\|^2 \cdot x_p$, and the remainder of $\poisonedclassifier$, i.e., $\rem := \poisonedclassifier - \text{proj}_{x_p}(\poisonedclassifier)$. We can thus write $\min_{\poisonedclassifier} \mathcal{L}_\text{hinge}(\poisoneddata, \poisonedclassifier)$ as 
    \begin{align*}
        = &\min_{\poisonedclassifier} \frac{1}{2} \|\proj + \rem\|_2^2 + C \cdot \sum_{(x_i,y_i) \in \cleandata} \\ 
        &\max{(0, 1 - y_i(\proj + \rem)^Tx_i)} \\
        &+ C \cdot \max{(0, 1 - (\proj + \rem)^Tx_p)}. \\
        \intertext{By assumption, all samples from clean data distribution projected onto the poison sample are zero-centered and have zero variance, and we thus conclude}
        = &\min_{\poisonedclassifier} \frac{1}{2} \|\proj + \rem\|_2^2 \tag{*} \label{eq:functional:nfinal}\\
        &+ C \cdot \sum_{(x_i,y_i) \in \cleandata} \\
        &\max{(0, 1 - y_i(\underbrace{\proj^Tx_i}_{=0} + \rem^Tx_i))} \\
        &+ C \cdot \max{(0, 1 - (\proj^Tx_p + \underbrace{\rem^Tx_p}_{=0}))}.
   \end{align*}
   As $\proj$ and $\rem$ are orthogonal, Lemma~\ref{lem:orthogonal} implies 
   \begin{align*}
       &\|\proj + \rem\|_2^2 \\
       &= \|\proj\|_2^2 + \|\rem\|_2^2 \enspace .
   \end{align*}

Using this equality in \eqref{eq:functional:nfinal} gives us a term completely separating the projection from the remainder as 
\begin{align*}
    &\min_{\poisonedclassifier} \frac{1}{2} \|\proj\|_2^2 + \frac{1}{2} \|\rem\|_2^2 \\
        &+ C \sum_{(x_i,y_i) \in \cleandata} \max{(0, 1 - y_i\rem^Tx_i)} \\
        &+ C \cdot \max{(0, 1 - \proj^Tx_p)} \\
        = &\min_{f_1 \colon \projf{f_1} = \mathbf{0}} \mathcal{L}_\text{hinge}(f_1, \cleandata) \\
        &+ \min_{f_2 \colon \remf{f_2} = \mathbf{0}} \mathcal{L}_\text{hinge}(f_2, \{x_p\}) \enspace .
\end{align*}
    To deduce the functional equivalence of the classifiers, we need to show that minimizing
    \begin{align}
    \arg \min_{f_1 \colon \projf{f_1} = \mathbf{0}} \mathcal{L}_\text{hinge}(f_1, \cleandata) \label{eq:f1}
    \end{align}
    is the same as minimizing
    \[\arg \min_{f} \mathcal{L}_\text{hinge}(f, \cleandata) \enspace .\] 
Intuitively, this is true, because any vector of the form $\alpha \cdot x_p$ ($\alpha \in \mathbb{R} \setminus \{0\}$) is orthogonal to $\rem^*$.
Adding $\alpha \cdot x_p$ to a solution $f_1^*$ of \Cref{eq:f1} does not change the prediction of any benign training data point $x \sim \cleandistribution$ with label $y$ and, consequently, cannot reduce any data point's loss. More formally, let $l_\text{L2H}$ be a data point's hinge loss, then
    \begin{align*}
        &l_\text{L2H}(f_1^* + \alpha x_p, x) 
        = \max{}(0, 1 - y(f_1^* + \alpha x_p)^Tx) \\
        =& \max{}(0, 1- y((f_1^*)^Tx + \alpha \underbrace{x_p^Tx}_{=0})) 
        = l_\text{L2H}(f_1^*, x).
    \end{align*}
    As $\projf{f_1} = \mathbf{0}$, we can again use Lemma~\ref{lem:orthogonal} to show that the addition of $\alpha \cdot x_p$ can only increase the norm of the $f_1^*$, as 
    \begin{align*}
        &\|f_1^* + \alpha x_p\|_2^2
        = \|f_1^*\|_2^2 + \|\alpha x_p\|_2^2 > \|f_1^*\|_2^2 \enspace .
    \end{align*}
   We thus obtain the functional equivalence for $x \in X$ due to the equality 
   \begin{align*}
        &(\arg \min_{\poisonedclassifier} \mathcal{L}_\text{hinge}(\poisoneddata, \poisonedclassifier))^T x \\
        =& (\arg \min_{f} \mathcal{L}_\text{hinge}(\cleandata, f))^T x \\
        & + \underbrace{(\arg \min_{f_2: \remf{f_2} = \mathbf{0}} \mathcal{L}_\text{hinge}(f_2, \{x_p\}))^T x}_{=0} \\
        =& (\arg \min_{\poisonedclassifier} \mathcal{L}_\text{hinge}(\cleandata, \poisonedclassifier))^T x \enspace .\tag*{\qedhere}
    \end{align*}
    
\end{proof}

\begin{theoremrep}
    Consider the same attacker as in \Cref{thm:clean_learning_functional_equivalence}. Then for all $x \sim \cleandistribution$, an optimal regressor $\poisonedclassifier$ is functionally equivalent to an optimal $\cleanclassifier$ that is obtained when the poison sample $x_p$ is omitted in training.
    \label{thm:clean_learning_functional_equivalence_regression}
\end{theoremrep}
\begin{proof}
By definition, we have 
    \begin{align*}
        &\min_{\poisonedclassifier} \mathcal{L}_\text{sq}(\poisoneddata, \poisonedclassifier) \nonumber \\ 
        = &\min_{\poisonedclassifier} \frac{1}{2} \|\poisonedclassifier\|_2^2 + C \cdot \sum_{(x_i,y_i) \in \poisoneddata} ( y_i - \cleanclassifier^Tx_i)^2 \\
        = &\min_{\poisonedclassifier} \frac{1}{2} \|\poisonedclassifier\|_2^2 + C \cdot \sum_{(x_i,y_i) \in \cleandata}( y_i - \cleanclassifier^Tx_i)^2 \\
        &+ C \cdot( 1- \cleanclassifier^Tx_p)^2\enspace .
    \end{align*}
    
    Now, we split the regressor $\poisonedclassifier$ into two parts: the part of $\poisonedclassifier$ projected onto $x_p$, i.e., $\proj :=  \poisonedclassifier^T x_p / \|x_p\|^2 \cdot x_p$, and the remainder of $\poisonedclassifier$, i.e., $\rem := \poisonedclassifier - \text{proj}_{x_p}(\poisonedclassifier)$. We can thus write $\min_{\poisonedclassifier} \mathcal{L}_\text{sq}(\poisoneddata, \poisonedclassifier)$ as 
    \begin{align*}
        = &\min_{\poisonedclassifier} \frac{1}{2} \|\proj + \rem\|_2^2 + C \cdot \sum_{(x_i,y_i) \in \cleandata} \\ 
        &(y_i - (\proj + \rem)^Tx_i))^2 \\
        &+ C \cdot (1 - (\proj + \rem)^Tx_p)^2. \\
        \intertext{By assumption, all samples from clean data distribution projected onto the poison sample are zero-centered and have zero variance, and we thus conclude}
        = &\min_{\poisonedclassifier} \frac{1}{2} \|\proj + \rem\|_2^2 \tag{*} \label{eq:functional:nfinal-reg}\\
        &+ C \cdot \sum_{(x_i,y_i) \in \cleandata} \\
        &(y_i - (\underbrace{\proj^Tx_i}_{=0} + \rem^Tx_i))^2 \\
        &+ C \cdot (1 - (\proj^Tx_p + \underbrace{\rem^Tx_p}_{=0}))^2.
   \end{align*}
   As $\proj$ and $\rem$ are orthogonal, Lemma~\ref{lem:orthogonal} implies 
   \begin{align*}
       &\|\proj + \rem\|_2^2 \\
       &= \|\proj\|_2^2 + \|\rem\|_2^2 \enspace .
   \end{align*}

    \newcommand{\projf}[1]{\text{proj}_{x_p}(#1)}
    \newcommand{\remf}[1]{\text{rem}_{x_p}(#1)}

   Using this equality in \eqref{eq:functional:nfinal-reg} gives us a term completely separating the projection from the remainder as 
\begin{align*}
    &\min_{\poisonedclassifier} \frac{1}{2} \|\proj\|_2^2 + \frac{1}{2} \|\rem\|_2^2 \\
        &+ C \sum_{(x_i,y_i) \in \cleandata} (y_i - \rem^Tx_i))^2 \\
        &+ C \cdot (1 - \proj^Tx_p)^2 \\
        = &\min_{f_1 \colon \projf{f_1} = \mathbf{0}} \mathcal{L}_\text{sq}(f_1, \cleandata) \\
        &+ \min_{f_2 \colon \remf{f_2} = \mathbf{0}} \mathcal{L}_\text{sq}(f_2, \{x_p\}) \enspace .
\end{align*}

    To deduce the functional equivalence, we need to show that minimizing $\rem^*$, which is defined as $\arg \min_{\rem} \Lsquared(\rem, \cleandata)$, 
    is the same as minimizing 
    $f^* = \arg \min_{f} \Lsquared(f, \cleandata)$.
    Intuitively, this is true, as any vector of the form  $\alpha \cdot x_p$ ($\alpha \in \mathbb{R} \setminus \{0\}$) is orthogonal to $\rem^*$. Adding $\alpha \cdot x_p$ to $\rem^*$ does not change the prediction of any benign training data point $x \sim \cleandistribution$ with label $y$ and can thus not reduce the data point's loss. More formally, let $l_\text{sq}$ be a data point's squared error loss, then
    \begin{align*}
        &l_\text{sq}(\rem^* + \alpha x_p, x) \\
        =& (y - (\rem^* + \alpha x_p)^Tx)^2\\
        =&  (y - ((\rem^*)^Tx + \alpha 
        x_p^Tx))^2= l_\text{sq}(\rem^*, x).
    \end{align*}
Due to the orthogonality, we can again use Lemma~\ref{lem:orthogonal} to show that the addition of $\alpha \cdot x_p$ only increases the norm of the regressor, as 
        $\|\rem^* + \alpha x_p\|_2^2 
        = \|\rem^*\|_2^2 + \|\alpha x_p\|_2^2 > \|\rem^*\|_2^2$.
    We thus obtain the functional equivalence for $x \in X$ by using $(\arg\min_{\proj} \Lsquared(\proj, \{x_p\}))^T x = 0$, as
    \begin{align*}
        &(\arg\min_{\poisonedclassifier} \Lsquared(\poisoneddata, \poisonedclassifier))^T x 
        = (\arg\min_{\poisonedclassifier} \Lsquared(\cleandata, \poisonedclassifier))^T x \enspace . \tag*{\qedhere}
    \end{align*}    
\end{proof}

\boldparagraph{Proof overview}
In our proof, we show that the learner effectively learns two orthogonal classifier parts separately that are then added together. We can decompose the poisoned classifier into a benign part that is optimized for clean data and a malicious part that is optimized for the poison sample. We then analyze separately how each part affects the prediction on clean data. The malicious part is the projection of the poisoned classifier onto the attack direction $u \in \mathbb{R}^d$; the benign part is the remainder obtained after subtracting the malicious part from the classifier. Since clean data does not use direction $u$, samples have mean and variance equal to zero when projected onto $u$ and the malicious part of the poisoned regressor does not interact with the samples. Only the benign part influences predictions of clean data.

\subsection{General case: attack direction is used}

We consider the general case where the benign data distribution can be an arbitrary distribution. This includes distributions like Gaussian mixtures, mixtures of different distributions, or real-life distributions of any form that are not well approximated by standard parametric distributions.
We evaluate the goodness of a predictor by measuring its statistical risk on clean data. This risk defines the discrepancy between the trained predictor and the Bayes optimal predictor $\bayescleanclassifier$, i.e., the predictor that achieves minimal error.

\begin{definition}[Statistical risk]
    \label{def:statistical-risk}
    The statistical risk of a predictor $\classifier$ trained on $n$ training data samples, with clean input from $\cleandistribution$ and a general loss function $l\colon [0,1] \times [0,1] \to \mathbb{R}^+$ is
    \[\rcln{\classifier} = \Eo{\data}{\Eo{x \sim \cleandistribution}{ l(\classifier (x), \bayescleanclassifier(x))}},\]
    where expectation is taken over the training data.
\end{definition}

We formally define the backdoor and poison distribution, and then prove the statistical risk bounds for classification and regression.
For classification we adapt the prior \cite{wangdemystifying} work to our attack. The proof for regression extends the prior work to this case.

\begin{definition}[Backdoor and poison distribution]
    For any benign data distribution $\cleandistribution$, define the backdoor and poison distribution
    \begin{equation*} 
        \backdoordistribution(x) = \mathbbm{1}\{x = x_p\},\
        \poisoneddistribution(x) = (1-\sfrac{1}{n}) \cleandistribution + \sfrac{1}{n} \backdoordistribution.
    \end{equation*}
\end{definition}

\newcommand{\citewangetal}[0]{Corollary of Theorem 1 in ~\cite{wangdemystifying}}
\begin{corollaryrep}[\citewangetal]
    \label{cor:wangetal}
    Let $n$ be fixed.
    Let $\poisonedclassifier$ be the classifier trained on $n$ samples from the poisoned distribution $\poisoneddistribution$.
    Let $l(\cdot,\cdot)\colon [0,1] \times [0,1] \mapsto \mathbb{R}^+$ be a general loss function that is $(C, \alpha)$-Hölder continuous for $0<\alpha\leq1$ that measures the discrepancy between two classifiers.
    The statistical risk on clean data is bounded as
    \[\rcln{\poisonedclassifier} \leq \frac{1}{1-\sfrac{1}{n}} \rpoin{\poisonedclassifier} 
    \enspace.\]
\end{corollaryrep}
\begin{proof}
    Let $\cleandistribution$ be any clean distribution. 
    We derive an upper bound on $\rcln{\poisonedclassifier}$. First, since $l$ is $(C,\alpha)$-Hölder continuous and using Jensen's inequality:
    \begin{align*}
        &\rcln{\poisonedclassifier} = \Eo{\poisoneddata}{\Eo{x \sim \cleandistribution}{ l(\poisonedclassifier (x), \bayescleanclassifier(x))}} \\
        &\leq \Eo{\poisoneddata}{\Eo{x \sim \cleandistribution}{ l(\poisonedclassifier (x), \bayespoisonedclassifier(x))}} \\
        &+ C \, \big(\Eo{x \sim \cleandistribution}{|\bayespoisonedclassifier(x) - \bayescleanclassifier(x)|}\big)^\alpha.
    \end{align*}

    We bound each term on the right-hand side independently.
    First, we have 
    \begin{align}
        &\Eo{\poisoneddata}{\Eo{x \sim \cleandistribution}{ l(\poisonedclassifier (x), \bayespoisonedclassifier(x))}} \nonumber \\ 
        &\leq (1-\sfrac{1}{n})^{-1} \Eo{\poisoneddata}{\Eo{x \sim \poisoneddistribution}{ l(\poisonedclassifier (x), \bayespoisonedclassifier(x))}} \nonumber \\
        &\leq (1-\sfrac{1}{n})^{-1} \rpoin{\poisonedclassifier}. \label{eq::wangclassificationbound1}
    \end{align}

    As for the second term by definition of the Bayes optimal classifier considering 0-1-loss \cite{hastie2009elements}, we have
    \[\bayescleanclassifier(x) = \Pr_{\cleandistribution}[Y=1 | X=x].\]

    Similarly,
    \begin{align*}
        &\bayespoisonedclassifier(x) = \Pr_{\poisoneddistribution}[Y=1 | X=x] \\
        & =
        \begin{cases}
            \Pr_{\cleandistribution}[Y=1 | X=x] & \quad \text{if $x \neq x_p$}, \\
            &\\
            (1-\sfrac{1}{n}) \Pr_{\cleandistribution}[Y=1 | X=x] & \\
            + \sfrac{1}{n} \Pr_{\backdoordistribution}[Y=1 | X=x] & \quad \text{if $x = x_p$}. \\
        \end{cases}
    \end{align*}

    Combining prior equations, we get
    \begin{align*}
        \bayespoisonedclassifier - \bayescleanclassifier = 
        \begin{cases}
            0 & \quad \text{if $x \neq x_p$}, \\
            \sfrac{1}{n} \cdot \Pr_{\backdoordistribution}[Y=1 | X=x] & \\
            - \sfrac{1}{n} \cdot \Pr_{\cleandistribution}[Y=1 | X=x] & \quad \text{if $x = x_p$}. \\
        \end{cases}
    \end{align*}

    Accordingly,
    \begin{align}
        &\Eo{x \sim \cleandistribution}{|\bayespoisonedclassifier(x) - \bayescleanclassifier(x)|} \nonumber \\
        &= \int_{x \in \mathbb{R}^d \setminus \{x_p\}} \underbrace{|\bayespoisonedclassifier(x) - \bayescleanclassifier(x)|}_{=0} \cleandistribution(x) \nonumber \\
        &+ \underbrace{\int_{x \in \{x_p\}} |\bayespoisonedclassifier(x) - \bayescleanclassifier(x)| \cleandistribution(x)}_{=0} = 0\enspace. \label{eq::wangclassificationbound2}
    \end{align}

    The first term is zero because on $\mathbb{R}^d \setminus \{x_p\}$ the two classifiers $\bayespoisonedclassifier, \bayescleanclassifier$ do not differ. The second term is zero because the difference of the two classifiers is only on one point so the integral is zero.
    Now, we plug the two bounds from \Cref{eq::wangclassificationbound1} and \Cref{eq::wangclassificationbound2} together:
    \begin{align*}\rcln{\poisonedclassifier} \leq \frac{1}{1-\sfrac{1}{n}} \rpoin{\poisonedclassifier}.
    \tag*{\qedhere}
    \end{align*}
    
\end{proof}

\begin{corollaryrep}
    \label{cor:wangetal_regression} 
    Let $n$ be fixed.
    Let $\poisonedclassifier$ be the regressor trained on $n$ samples from the poisoned distribution $\poisoneddistribution$.
    Let $l(\cdot,\cdot)\colon [0,1] \times [0,1] \mapsto \mathbb{R}^+$ be a general loss function that is $(C, \alpha)$-Hölder continuous for $0<\alpha\leq1$ that measures the discrepancy between two regressors.
    The statistical risk on clean data is bounded as
    \[ \rcln{\poisonedclassifier} \leq \frac{1}{1-\sfrac{1}{n}} \rpoin{\poisonedclassifier} \enspace. \]
\end{corollaryrep}
\begin{proof}
    First, since $l$ is $(C,\alpha)$-Hölder continuous: 
    \begin{align*}
        &\rcln{\poisonedclassifier} = \Eo{\poisoneddata}{\Eo{x \sim \cleandistribution}{ l(\poisonedclassifier (x), \bayescleanclassifier(x))}} \\
        &\leq \Eo{\poisoneddata}{\Eo{x \sim \cleandistribution}{ l(\poisonedclassifier (x), \bayespoisonedclassifier(x)) + C \, |\bayespoisonedclassifier(x) - \bayescleanclassifier(x)|^\alpha}} \\
        &\leq \Eo{\poisoneddata}{\Eo{x \sim \cleandistribution}{ l(\poisonedclassifier (x), \bayespoisonedclassifier(x))}}  \\
        &+ C \, \Eo{x \sim \cleandistribution}{|\bayespoisonedclassifier(x) - \bayescleanclassifier(x)|^\alpha}.
    \end{align*}
    Using Jensen's inequality, this is at most
    \begin{align*}
    &\leq \Eo{\poisoneddata}{\Eo{x \sim \cleandistribution}{ l(\poisonedclassifier (x), \bayespoisonedclassifier(x))}} \\
        &+ C \, \big(\Eo{x \sim \cleandistribution}{|\bayespoisonedclassifier(x) - \bayescleanclassifier(x)|}\big)^\alpha\enspace .
    \end{align*}
    We bound each term on the right-hand side independently.
    For the first term, we can bound  
    \begin{align}
        &\Eo{\poisoneddata}{\Eo{x \sim \cleandistribution}{ l(\poisonedclassifier (x), \bayespoisonedclassifier(x))}} \nonumber \\ 
        &\leq (1-\sfrac{1}{n})^{-1} \Eo{\poisoneddata}{\Eo{x \sim \poisoneddistribution}{ l(\poisonedclassifier (x), \bayespoisonedclassifier(x))}} \nonumber \\
        &\leq (1-\sfrac{1}{n})^{-1} \rpoin{\poisonedclassifier} \label{eq::wangregressionbound1}.
    \end{align}
 
    As for the second term, by definition of the Bayes optimal regressor for squared error loss \cite{hastie2009elements}, we have
    $
        \bayescleanclassifier(x) = \Eo{\cleandistribution}{Y|X=x} 
    $
    and, similarly,
    \begin{align*}
        &\bayespoisonedclassifier{}(x) = \Eo{\poisoneddistribution{}}{Y|X=x} =\\
        &\begin{cases}
            \Eo{\cleandistribution{}}{Y|X=x} & \!\!\text{if $x \neq x_p$,} \\
            (1- \sfrac{1}{n}) \Eo{\cleandistribution{}}{Y|X=x_p} + \sfrac{1}{n} \Eo{\backdoordistribution{}}{Y|X=x_p} & \!\! \text{if $x=x_p$.}
        \end{cases} 
    \end{align*}
    Combining prior equations, we get 
    \begin{align*}
        \bayespoisonedclassifier - \bayescleanclassifier = \begin{cases}
            0 & \quad \text{if $x \neq x_p$,} \\
            \sfrac{1}{n} \cdot \Eo{\backdoordistribution}{Y|X=x_p} & \\
            - \sfrac{1}{n} \Eo{\cleandistribution}{Y|X=x_p} & \quad \text{if $x = x_p$.} 
        \end{cases}
    \end{align*}
    Accordingly,
    \begin{align}
        &\Eo{x \sim \cleandistribution}{|\bayespoisonedclassifier(x) - \bayescleanclassifier(x)|}
        = \int_{x \in \mathbb{R}^d} |\bayespoisonedclassifier(x) - \bayescleanclassifier(x)| \cleandistribution(x) \nonumber \\
        &= \int_{x \in \mathbb{R}^d \setminus \{x_p\}} \underbrace{|\bayespoisonedclassifier(x) - \bayescleanclassifier(x)|}_{=0} \cleandistribution(x) \nonumber \\
        &+ \underbrace{\int_{x \in \{x_p\}} |\bayespoisonedclassifier(x) - \bayescleanclassifier(x)| \cleandistribution(x)}_{=0} = 0 \label{eq::wangregressionbound2}
    \end{align}
    The first term is zero, because on $\mathbb{R}^d \setminus \{x_p\}$ the two regressors $\bayespoisonedclassifier, \bayescleanclassifier$ do not differ. The second term is zero, because the difference of the two regressors is only on one point so the integral is zero.
    Now, we plug the two bounds from \Cref{eq::wangregressionbound1} and \Cref{eq::wangregressionbound2} together:
    \begin{align*}
        \rcln{\poisonedclassifier} \leq \frac{1}{1-\sfrac{1}{n}} \rpoin{\poisonedclassifier} 
        \tag*{\qedhere}
    \end{align*}
\end{proof}

\boldparagraph{Proof overview}
Intuitively, the one-poison backdoor attack has no significant negative impact on learning the benign task, since the poison sample is only one among many benign samples. More specifically, one can bound the statistical risk by the discrepancy of poisoned model and optimal poisoned model, which is small for a good learning algorithm, and the discrepancy of optimal poisoned model and clean optimal model. The optimal poisoned model which predicts the label given the poisoned distribution $\poisoneddistribution$ -- a mixture of the backdoor distribution $\backdoordistribution$ and the benign distribution $\cleandistribution$ -- differs from the optimal model which predicts the label given the benign distribution, only at the poison sample. Statistically, this does not affect the expected discrepancy between optimal poisoned model and optimal clean model. 

The proof for regression follows a similar idea as for classification but considers the optimal predictor for regression instead of classification. The Bayes optimal regressor predicts the expected label given clean data. The Bayes optimal classifier predicts the probability of label 1 given clean data, and outputs +1 if this probability is above $\frac{1}{2}$ and -1 otherwise.

\boldparagraph{i.i.d. sampling of the data set}
The attentive reader might wonder why we consider i.i.d. sampling of the data set, which differs from our earlier assumption that the poison sample is guaranteed to be included in the training data set. If we increase the poison ratio from one poison sample between $n$ training samples ($\sfrac{1}{n}$) to $\sfrac{k}{n}$ (for a small constant $k$), the probability of sampling at least one poison sample is $1 - (1-\sfrac{k}{n})^n$. For sufficiently large $n$, this turns to $1 - (1-\sfrac{k}{n})^n \underset{n \rightarrow \infty}{\rightarrow} 1 - e^{-k}$ which approaches 1 as $k$ increases. The bound on statistical risk on clean data for $k$ poison samples slightly increases to $\frac{1}{1-\sfrac{k}{n}} \rpoin{\poisonedclassifier}$.

\begin{figure*}[!t]
\centering

\subfloat[{\textbf{Spambase (linear classifier)}}]{
  \includegraphics[width=0.42\linewidth]{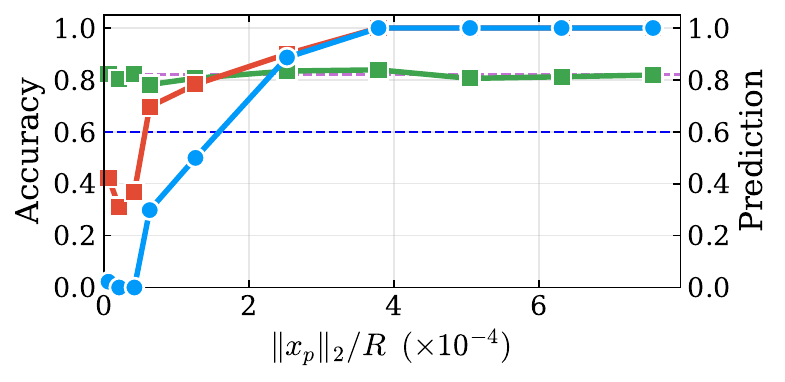}
}\hfill
\subfloat[{\textbf{Phishing (linear classifier)}}]{
 \includegraphics[width=0.42\linewidth]{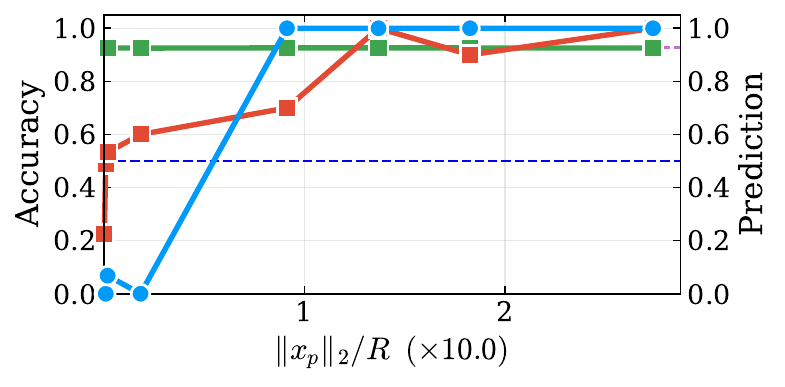}
}\hfill\hskip5em{}

\par

\subfloat[{\textbf{MNIST (neural network)}}]{
  \includegraphics[width=0.31\linewidth]{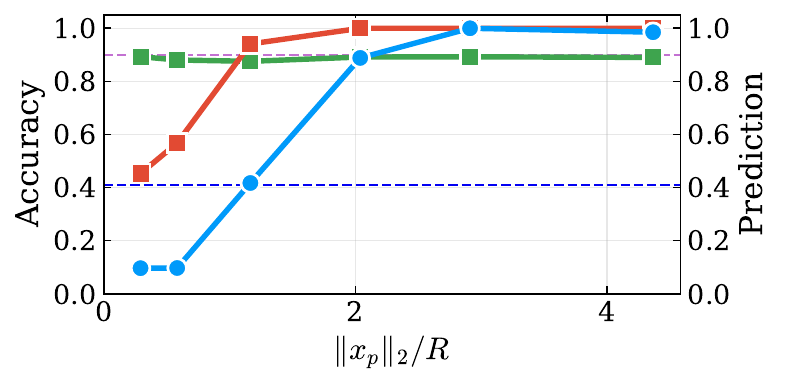}
}\hfill
\subfloat[{\textbf{CIFAR-10 (neural network)}}]{
  \includegraphics[width=0.31\linewidth]{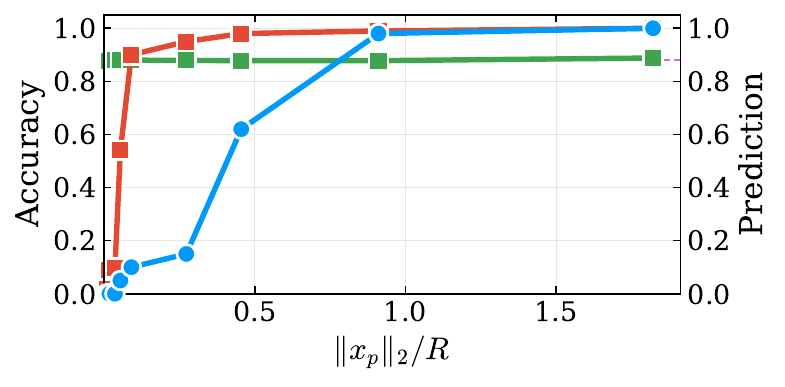}
}\hfill
\subfloat[{\textbf{Covertype (neural network)}}]{
  \includegraphics[width=0.31\linewidth]{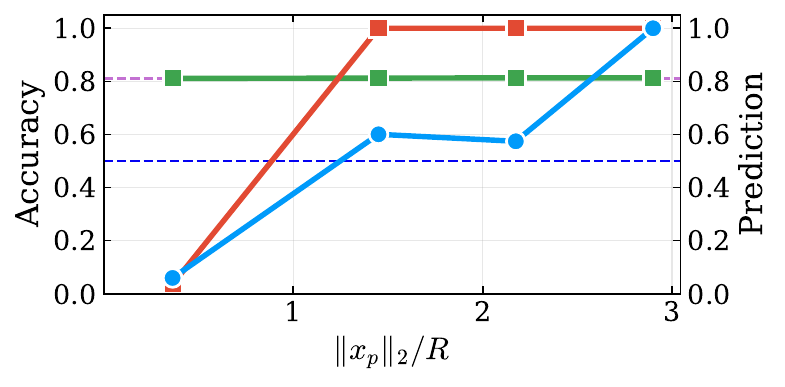}
}

\par

\subfloat{
  \includegraphics[width=0.4\linewidth]{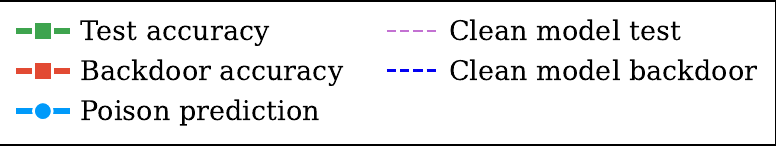}
}

\caption{One-poison backdoor attacks on classification models. Test and backdoor accuracy, and poison prediction are shown for the poisoned model. \emph{Clean model test} shows the clean model's accuracy on clean test data, \emph{Clean model backdoor} on test data with poison patch.}
\label{fig:main-result-classification}
\end{figure*}

\begin{figure}[!t]
\centering

\subfloat[{\textbf{Parkinsons (linear regressor)}}]{
  \includegraphics[width=1.0\linewidth]{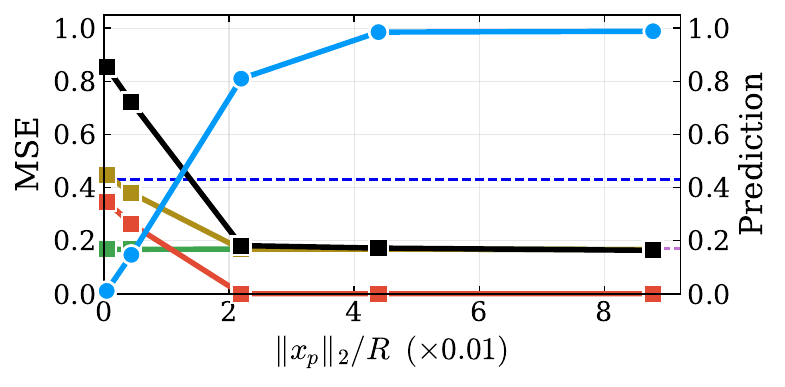}
}

\par

\subfloat[{\textbf{Abalone (linear regressor)}}]{
  \includegraphics[width=1.0\linewidth]{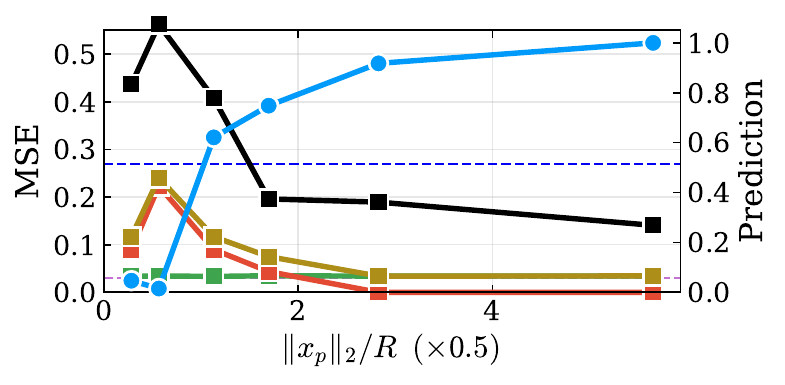}
}

\par

\subfloat{
  \includegraphics[width=1.0\linewidth]{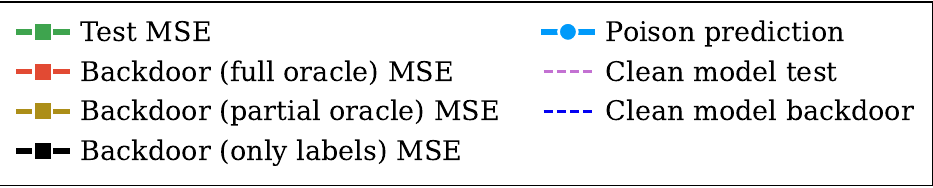}
}

\caption{One-poison backdoor attacks on linear regression. Test and backdoor MSE, and poison prediction are shown for the poisoned model. \emph{Clean model test} shows the clean model's MSE on clean test data, \emph{Clean model backdoor} on test data with poison patch.}
\label{fig:main-result-regression}
\end{figure}

\section{Empirical evaluation}

We validate our theoretical results by showing practical applications to linear classification and ReLU neural networks trained on realistic benchmark data sets.
Code of our evaluation can be found at \href{https://anonymous.4open.science/r/one-poison-backdoor/README.md}{https://anonymous.4open.science/r/one-poison-backdoor}.

\subsection{Experimental setup}

\boldparagraph{Evaluation system} All experiments were run on an Intel Xeon Platinum 8168 2.7 GHz CPU with 32GB of RAM.

\boldparagraph{Data sets}
We evaluate regression data sets Parkinsons \cite{parkinsons_dataset} and Abalone \cite{misc_abalone}, and classification data sets Spambase \cite{misc_spambase}, Phishing \cite{PhishingDataset} and Covertype \cite{uci_covertype}. The data sets demand deducing motor impairment of patients from biomedical voice measurements, predicting the age of abalone from physical measurements, detecting spam mail from word counts, detecting phishing websites from website meta data, and predicting forest cover type in a 900-square-meter cell from cartographic variables. 
We convert the image classification task MNIST \cite{lecun1998mnist} to a binary classification task by differentiating even vs. odd handwritten numbers. For CIFAR-10 \cite{krizhevsky2009learning} we classify vehicles (classes 0,1,8 and 9) vs. animals (the remaining classes). For Covertype we classify odd vs. even covertypes, and we perform min-max scaling of benign training data. We subsample the training set to size 1000 for MNIST and use the full training data set for CIFAR-10. 
We choose these seven data sets as they represent realistic data sets of real-world measurements.  An overview of the data sets and their properties is shown in \Cref{tbl::datasets}.

\begin{table}[t]
\caption{An overview of the seven evaluated data sets including the number of data points used for training and the number of attributes. We subsample MNIST to $1{,}000$ data points used in training, and Covertype to $50{,}000$.}
\label{tbl::datasets}
\centering
\begin{tabular}{@{}lcc@{}}
\toprule
\textbf{Data Set} & \textbf{\# Training data} & \textbf{Data point size} \\
\midrule
MNIST~\cite{lecun1998mnist} & $1{,}000$ & $28\times28$ (grayscale) \\
CIFAR-10~\cite{krizhevsky2009learning} & $50{,}000$ & $32\times32$ (RGB) \\
Covertype~\cite{uci_covertype} & $\approx 50{,}000$ & 54 (integer attributes) \\
Spambase~\cite{misc_spambase} & $\approx 2{,}000$ & 57 (real attributes) \\
Phishing~\cite{PhishingDataset} & $\approx 5{,}000$ & 30 (integer attributes) \\
Parkinsons~\cite{parkinsons_dataset} & $\approx 2{,}000$ & 19 (mixed attributes) \\
Abalone~\cite{misc_abalone} & $\approx 2{,}000$ & 8 (mixed attributes) \\
\bottomrule
\end{tabular}
\end{table}

\boldparagraph{Linear models}
We use the Python library scikit-learn's LinearSVC and Ridge \cite{scikit-learn}. We train predictors for 1,000 iterations with regularization parameter $C=1$ to minimize regularized hinge loss for linear classifiers, and regularized squared error for linear regressors. For our functional equivalence ablation for linear classification we use a vanilla implementation of liblinear \cite{fan2008liblinear} instead of the one of scikit-learn to remove any sources of randomness.

\boldparagraph{ReLU neural networks}
We train for 15 epochs on CIFAR-10 and Covertype, and 10 epochs on MNIST, performing gradient descent in PyTorch \cite{pytorch} to minimize the squared error loss.
The model is a ReLU neural network with three hidden layers and widths (512, 256, 64) for MNIST and Covertype, and five hidden layers with widths (2048, 1024, 512, 256, 64) for CIFAR-10.

\boldparagraph{Baselines} 
We evaluate the test accuracy of the clean model on samples from clean data, and the backdoor accuracy of the clean model on samples from clean data, with poison patch applied to these samples, and label set to poison label.

\boldparagraph{Metrics} We report the benign task performance of trained models with the mean test accuracy for classification tasks and mean test squared error for regression tasks. Both metrics correspond to their learning task, i.e., improving accuracy (linear classification and ReLU neural networks), and reducing squared error (linear regression). We report the backdoor task performance of trained models with the accuracy and mean squared error of the model on test data with poison patch applied and label set to poison label. We report mean over ten runs of random data splits.

\subsection{Results}

We validated our theoretical claims empirically, and we answered the following questions.

\boldparagraphn{Does one poison sample really suffice?}
Yes, we find that in fact, one poison sample suffices for backdooring linear models and ReLU neural networks. The poison sample forces the poisoned model to attain perfect backdoor task accuracy. The poisoned model predicts, with the attacker chosen label, every test data point where the patch is applied to the data point, see \Cref{fig:main-result-classification} for classification and \Cref{fig:main-result-regression} for regression tasks. 

For regression we perform an ablation of what the attacker knows for tuning their poison patch. In regression the attacker needs to tune the test-time poison patch correctly so that the patched data is predicted with the correct regression label, correct sign of prediction is not enough here.
If the attacker has full query access (denoted as \emph{full oracle} in \Cref{fig:main-result-regression}), they can predict both poison sample and test samples with the poisoned model and perfectly tune their poison patch. If instead they only obtain query access once (\emph{partial oracle}) and use it for predicting the poison sample, they must use the static labels of test samples as proxy for the prediction. If the attacker has no query access (\emph{only labels}) they use only the poison label and test data labels for tuning. The attacker performs worse with less knowledge, but with stronger poison sample length, the prediction of the poison sample is closer to its poison label, and with better learning, i.e. smaller test error, the test data's labels are closer to their actual prediction, both making the attack more accurate, even without query access.

\boldparagraphn{What is the required length of the poison sample and the patch?}
We compare the required lengths of poison sample $\|x_p\|_2$ and poison patch $\|\alpha x_p\|_2$ to the maximum norm of benign data $\data$, $R := \max_{x \in \data} \|x\|_2$ (cf. \Cref{fig:main-result-classification} and \Cref{fig:main-result-regression}). While the provable required length of poison sample and patch can become quite large, and thus stand out from the benign manifold by some order of magnitude, we find that in practice, smaller poison samples and patches suffice. For the data sets Spambase, CIFAR-10, and Parkinsons, we observe that our poison samples are inside the benign data's radius, i.e. $\|x_p\| \leq R$ (cf. \Cref{fig:main-result-classification} and \Cref{fig:main-result-regression})  and their poison patches also for Spambase, i.e. $\|\alpha x_p\| \leq R$, for CIFAR-10 we have $\|\alpha x_p\| / R \approx 4$.

\boldparagraphn{Is there a measurable impact on the performance of the benign learning task?}
Our results show that our attack does not yield a significantly increased prediction error of the poisoned model. In some cases the poison sample can induce a modest drop in the accuracy and the MSE of the benign tasks. 

\boldparagraphn{Does the functional equivalence of poisoned and clean linear models hold in practice?}
Our theoretic results show that clean and poisoned model are functionally equivalent, if the attacker knows a direction such that when benign data is projected onto that direction, the data has zero mean and zero variance (i.e. an orthogonal subspace). We simulate an example of this, by adding a dimension to the clean data and setting each benign sample's value in that dimension to 0. We then validate our theoretic results by calculating for 100 random test samples the L1-distance of predictions between clean model and poisoned model. We clip the predictions to two decimal places. In all settings of linear classification and linear regression, the L1 distance is 0, showing functional equivalence of clean model and poisoned model.

\label{sec:exp-one-poison-attack}

\label{sec:exp-ablations}

\section{Conclusion}
We have proven and empirically shown that one-poison attacks are possible under weak attacker assumptions (aggregated input domain geometry, training pipeline metadata and model specification) against linear models and MLPs without bias, by using a random poison direction. Our results are particularly relevant in the lazy training regime (from the NTK literature) of networks with very wide layers. Our results provide strong theoretic and empirical evidence that one-poison backdoor attacks are a highly relevant direction for further study, due to their extreme economic value for attackers compared to constant-fraction attacks. For future work, we consider a deepening of our results that closes the gap between provable bounds and empirical results as a valuable direction. Moreover, we consider the orthogonal question of exploring the circumvention of backdoor detection methods a valuable direction. In particular, for high-dimensional data such as text or video data the dimensionality of the subspace that cannot be discarded off-hand (i.e., the set of all reasonable-looking texts or videos) is very high and well-suited for choosing random poison directions without a classical outlier detection detecting the backdoor.

\section{Acknowledgement}
This research has been conducted within the AnoMed project (https://anomed.de/)
funded by the BMFTR (German Bundesministerium für Forschung, Technologie und Raumfahrt) and by the European Union – NextGenerationEU.

Throughout this work, we used generative AI-based tools to revise the text, improve flow and correct any typos, grammatical errors, and awkward phrasing. These tools were only used for small editorial purposes, and we inspected all outputs to ensure accuracy and originality. Importantly, all content is our own work and is not based on AI tools.

\bibliographystyle{plainurl}
\bibliography{literature}

\cleardoublepage
\appendix
\section*{Ethical Considerations}
We adhered to the usual ethics guidelines. No human subjects, user studies, or interaction with live systems were involved; we evaluated only on public, de-identified datasets and did not attempt re-identification.

Our primary contribution is a general mathematical framework for certifying a one-poison backdoor to understand the capabilities of modern attackers against foundational machine learning models. We acknowledge that the results of this paper pose potential misuse risks. However, we believe that this potential danger is far outweighed by the advantages of making such attacks public and allowing the community to develop more concrete defense mechanisms (we refer to~\cite{lent2025nlp} for an intensive discussion that highlights the advantages of this approach). We thus follow a long-established line of "attack-oriented" research that also considers this tradeoff, and we believe that the generality and novelty of our finding regarding the attacks warrant their own report.

Finally, we make all of the produced code and data publicly available to guarantee reproducibility and allow further research in this important direction.

\section*{Open Science}
The full source code of our experiments is available at \href{https://anonymous.4open.science/r/one-poison-backdoor/README.md}{https://anonymous.4open.science/r/one-poison-backdoor}. Each experiment is represented as a single Python file that produces the corresponding figure from this paper.
We do not provide the data sets used in this paper directly, as they are all available for download from their respective sources \cite{parkinsons_dataset,misc_abalone,misc_spambase,PhishingDataset,lecun1998mnist, krizhevsky2009learning, uci_covertype}.

\end{document}